\documentclass[sigconf]{acmart}

\AtBeginDocument{%
}

\usepackage{algorithm}
\usepackage{algorithmic}
\usepackage{graphicx}
\usepackage{caption}
\usepackage{subcaption}
\usepackage{booktabs}
\usepackage{multirow}

\newtheorem{definition}{Definition}

\newtheorem{example}{Example}

\newcommand{\myhead}[1]{\noindent\textbf{#1. }}

\newcommand{\methodname}{DeMix}
\newcommand{\eat}[1]{}
\newcommand{\camera}[1]{\textcolor{black}{#1}}

\copyrightyear{2026} 
\acmYear{2026} 
\setcopyright{cc} 
\setcctype{by-nc-nd}
\acmConference[KDD '26]{Proceedings of the 32nd ACM SIGKDD Conference on Knowledge Discovery and Data Mining V.2}{August 09--13, 2026}{Jeju Island, Republic of Korea} 
\acmBooktitle{Proceedings of the 32nd ACM SIGKDD Conference on Knowledge Discovery and Data Mining V.2 (KDD '26), August 09--13, 2026, Jeju Island, Republic of Korea} 
\acmDOI{10.1145/3770855.3817774} 
\acmISBN{979-8-4007-2259-2/2026/08} 

\begin{document}
\title{\methodname: Debugging Training Data with Mixed Data Error Types by Investigating Influence Vectors}

\author{Jiale Deng}
\authornote{Work done while Jiale Deng was an intern at ByteDance Inc.}
\affiliation{%
    \institution{Shanghai Jiao Tong University}
    \city{Shanghai}
    \country{China} 
}
\email{jialedeng@sjtu.edu.cn}

\author{Yanyan Shen}
\authornote{Corresponding author.}
\affiliation{%
    \institution{Shanghai Jiao Tong University}
    \city{Shanghai}
    \country{China} 
}
\email{shenyy@sjtu.edu.cn}

\author{Xiaogang Shi}
\affiliation{%
    \institution{ByteDance Inc.}
    \city{Beijing}
    \country{China} 
}
\email{shixiaogang@bytedance.com}

\author{Junjun Chai}
\affiliation{%
    \institution{Tiktok}
    \city{San Jose} \state{CA}
    \country{USA} 
}
\email{chaijunjun@bytedance.com}

\renewcommand{\shortauthors}{Jiale Deng, Yanyan Shen, Xiaogang Shi and Chai Junjun}

\begin{abstract}
High-quality training data is essential for the success of machine learning models. However, real-world datasets often contain mixed types of errors arising from systematic flaws in data preparation pipelines, including label errors, feature errors, and spurious correlations. Effective debugging of training data requires both detecting erroneous samples and identifying their specific error types to enable targeted repair, yet existing data cleaning and attribution methods fail to adequately address this dual requirement. In this paper, we propose \methodname, a novel framework that simultaneously diagnoses erroneous samples and their error types. Our key insight is that different error types produce distinct patterns on model behavior. \methodname~captures such error-specific patterns by influence vectors that characterize how each training sample affects model predictions across all validation samples. We formulate training data debugging as a multi-label classification problem where a classifier is developed to predict error types directly from influence vectors. We further introduce an intervention-based learning strategy that guides the classifier to capture invariant rationales specific to each error type, ensuring the learned classifier generalizes effectively. Empirical evaluations on 11 tasks across tabular data prediction, recommendation systems, and LLM alignment demonstrate that \methodname~significantly outperforms state-of-the-art approaches, achieving a 22.61\% improvement in data debugging F1-score and a 9.32\% gain in task model performance after data repair. Code is available at: \url{https://github.com/SJTU-DMTai/DeMix}.
\end{abstract}



\maketitle

\newcommand\kddavailabilityurl{https://doi.org/10.5281/zenodo.20502263}
\ifdefempty{\kddavailabilityurl}{}{
\begingroup\small\noindent\raggedright\textbf{Resource Availability:}\\
The source code of this paper has been made publicly available at \url{\kddavailabilityurl}.
\endgroup
}

\section{Introduction}

Data serves as the fundamental resource powering diverse machine learning applications, from recommendation systems~\cite{Dataset_regeneration_KDD24, Shapley_data_pruning_KDD25, Shapley_Recsys25} to large language model-based applications~\cite{LESS_ICML24, DataMan_ICLR25, Data_Attribution_Survey25, Influence_context_selection_NIPS25, Data_quality_FCS, Alignment_FCS}.
Training data quality has emerged as the primary determinant of model performance, establishing a new data-centric paradigm in machine learning deployment~\cite{Trustworthy_AI_Data_NatMachIntell22}. 
However, preparing data for model training involves a multi-stage process that typically encompasses data collection, transformation, feature engineering, and labeling~\cite{Trustworthy_AI_Data_NatMachIntell22}. 
Each stage may contain systematic flaws that introduce different types of errors into the final training dataset. 
Common error types include mislabeled samples caused by ambiguous annotation guidelines~\cite{Data_Glitches_KDD25, Disdetect_VLDB24, Influence_Relabeling_ICLR21}, corrupted features resulting from bugs in feature processing systems~\cite{Data_Glitches_KDD25, UniClean_VLDB25}, and spurious correlations arising from selection bias or confounding variables~\cite{Spurious_Correlation_KDD25, Discover_and_Cure_ICML23, ODIN_ICML24, Imbalance_FCS}. 
When models are trained on data containing such mixed error types, they inevitably learn erroneous and biased patterns, resulting in unreliable predictions and significant deployment risks.
Therefore, \emph{training data debugging} has become a crucial problem, which requires addressing two interconnected questions: \emph{which training samples are erroneous}, and \emph{what type of error do they contain}. 
Answering both questions jointly provides essential implications for locating systematic flaws and fixing them at their source.


Various efforts have been devoted to improving training data quality, which can be generally classified into two categories. 
Data cleaning methods~\cite{Data_Cleaning_Survey_SIGMOD16, SAGA_SIGMOD23, UniClean_VLDB25, Rock_Data_Cleaning_SIGMOD24} primarily assume erroneous samples exhibit statistical deviations from clean data and can thus be flagged via distributional analysis, outlier detection, or consistency checks.
While effective for random errors and isolated anomalies, these methods are limited in identifying systematic errors in training data. 
For instance, if a specific subgroup is consistently mislabeled due to incorrect labeling functions, none of them appear as anomalies relative to others in the same group, making data cleaning methods ineffective.
Recently, data attribution methods~\cite{Data_Attribution_Survey25, Influence_Analysis_Survey_24, Data_Glitches_KDD25, Shapley_Recsys25, Shapley_data_pruning_KDD25} have emerged as a promising alternative. They use influence functions~\cite{influence_ICML17, Influence_Relabeling_ICLR21, Influence_Analysis_Survey_24} to quantify how removing a training sample affects model performance on a validation set, flagging those that negatively impact the model as erroneous.
However, they mainly focus on identifying erroneous samples while leaving error type classification unsolved. Additionally, accurate data attribution relies on the availability of clean and unbiased validation sets, which are difficult to obtain in many practical applications such as recommendation systems where validation data often suffers from the same systematic errors as the training data.
To these ends, existing methods are inadequate for comprehensive training data debugging.

\begin{figure}[t]
  \centering
  \includegraphics[width=1\linewidth]{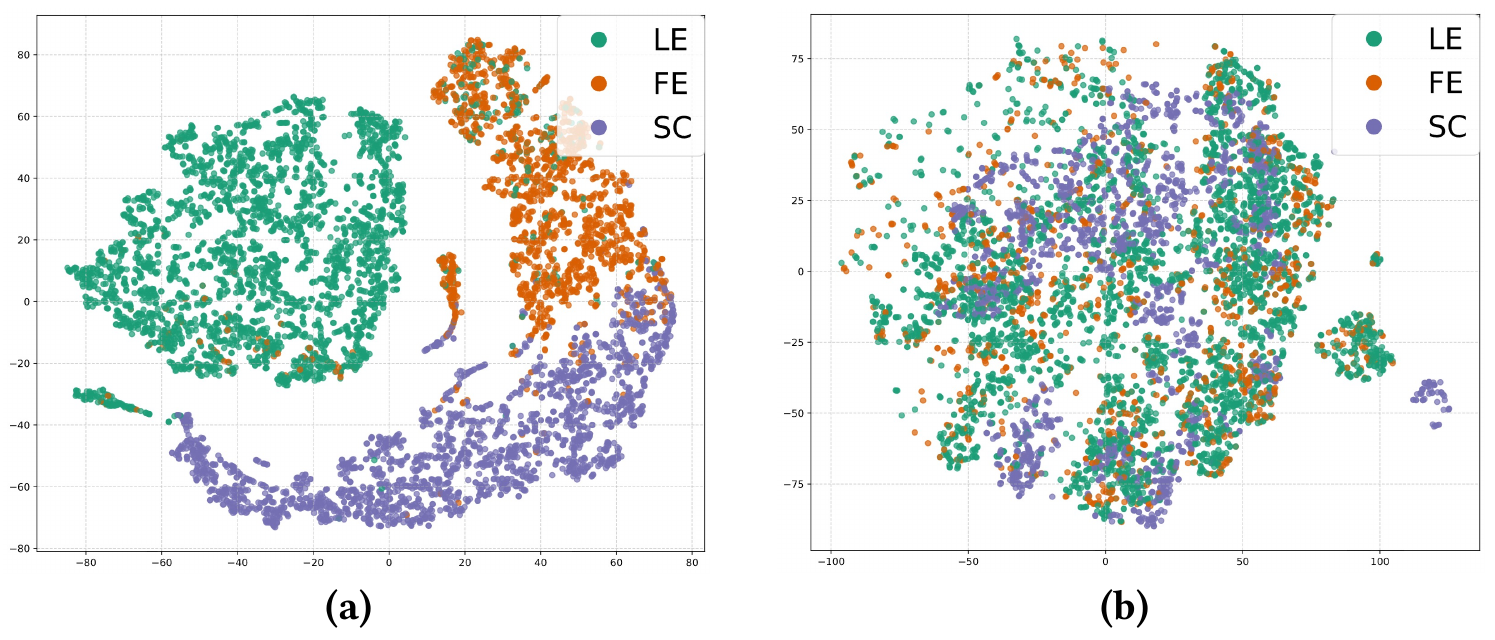}
  \caption{t-SNE visualization of (a) influence vectors and (b) raw features of erroneous samples, with three types of errors injected into the Adult dataset (more results in Appendix~\ref{app:visual}).}
  \vspace{-.1in}
  \label{fig:tsne}
\end{figure}

To address the problem, our key insight is that systematic training data errors often introduce consistent biases into model behavior, and these biases are reflected in how training samples influence model predictions across the entire validation set. Critically, different error types induce qualitatively different patterns on the validation data through influence function.
For example, samples with label errors tend to exert negative influence on validation samples with similar features but correct labels. In contrast, samples with spurious correlations often positively influence validation samples sharing spurious attributes while negatively influencing counterexamples that violate the spurious pattern.
Hence, we capture such error-specific patterns by using the complete influence vector rather than aggregating influence into a single scalar.
%
Formally, for a training sample $z_i$, the \emph{influence vector} is defined as $\Phi_i=[\phi_{i,1}, \dots, \phi_{i,M}]$, where $M$ is the validation set size. Each entry $\phi_{i,j}$ measures the impact of removing $z_i$ w.r.t. the loss of validation sample $z_j$, computed via influence functions.
Fig.~\ref{fig:tsne} provides empirical evidence of the distinguishing power of influence vectors: for a training sample $\{z_i\}$, we compare the t-SNE embeddings of its influence vector $\Phi_i$ and the raw features $\{z_i=(x_i, y_i)\}$.
Across the three error types including label errors (LE), feature errors (FE), and spurious correlations (SC), the visualization shows that influence vectors successfully disentangle error-specific clusters that remain mixed in the original data space. 

Based on our insight, we develop a multi-label classifier that takes the influence vector $\Phi_i$ of each training sample $z_i \in \mathcal{D}_{t}$ as input and predicts a set of error types denoted as $\hat{\rm t}_i$.
Since influence vectors encode interactions with an unordered validation set, we employ a Set Transformer~\cite{Set_transformer} to encode $\Phi_i$ into a low-dimensional representation, which is then decoded by multiple MLP heads for the final prediction. 
To facilitate supervised training of the classifier, we provide a controlled error injection strategy that generates synthetic datasets where selected training samples are deliberately corrupted and annotated with known error types.
Note that we do not assume that the validation set is perfectly clean or unbiased. Since we focus on identifying characteristic patterns rather than relying on absolute values in the influence vectors, our method can tolerate noisy validation data as long as the patterns induced by different error types remain distinguishable.

However, a key challenge in the proposed solution remains: influence vectors depend not only on the training data itself, but also on the configuration used for influence computation. 
Both the choice of the validation set and the task model instance affect the resulting vectors, even when the underlying training data remains unchanged.
Without proper controls, the classifier may exploit configuration-specific patterns that fail to generalize beyond the training setup. 
To mitigate this issue, we introduce an invariant representation learning strategy that encourages the Set Transformer encoder to extract error-specific patterns stable across different influence computation settings.
\camera{From an Information Bottleneck perspective~\cite{Information_bottleneck, DeepVIB_ICLR17, GSAT_ICML22}, this strategy learns a minimal sufficient representation that preserves error-type semantics while filtering configuration-specific noise (detailed analysis in Appendix~\ref{app:theory}).}
This is achieved through two kinds of interventions.
First, we intervene on the validation set by computing multiple influence vectors with different randomly sampled validation subsets, and apply a contrastive loss to keep their representations close.
Second, we intervene on the task model by computing influence vectors with an ensemble of models that differ in architecture or initialization, and align their representations via a pairwise consistency loss. 
Together, these losses force the encoder to focus on patterns that persist across configurations, thus improving its generalisability.

In this paper, we present \textbf{\methodname}, an automated framework designed for training data \textbf{De}bugging with \textbf{Mix}ed error types. The framework operates by first computing influence vectors for training samples and then feeding them into the classifier to predict erroneous samples along with their error types.
We evaluate \methodname~on 11 tasks spanning tabular data prediction, recommendation systems, and LLM alignment. The results demonstrate that \methodname~significantly outperforms state-of-the-art baselines, achieving a 22.61\% improvement in error type classification F1-score and a 9.32\% gain in the task model performance with repaired training data.


The main contributions of this paper are summarized as follows.
\begin{itemize}
    \item We investigate the under-explored problem of debugging training data with mixed error types, which requires identifying both erroneous samples and their corresponding error types. We reveal that influence vectors effectively capture the distinct patterns through which different error types affect model predictions.
    \item We propose \methodname, a novel framework that simultaneously diagnoses erroneous samples and their error types by formulating the problem as multi-label classification with influence vectors as inputs. We further introduce intervention-based training objectives that guide the classifier to capture invariant and error-specific patterns.
    \item Extensive experiments across diverse machine learning tasks and task models demonstrate that \methodname~outperforms state-of-the-art baselines by accurately debugging data with mixed error types and improving task model performance through targeted repair tailored to different error types.
\end{itemize}

\section{Preliminaries}
\label{section:ML}We consider a standard supervised ML setting involving a task model $f_\theta: \mathcal{X}\to \mathcal{Y}$ that maps an input $x_i$ from the input space $\mathcal{X}$ to a label $y_i$ in the label space $\mathcal{Y}$, where $\theta$ is the model parameters. Given a training dataset $\mathcal{D}_{t}=\{ z_i=(x_i,y_i) \}_{i=1}^N$ consisting of $N$ samples, the objective is to learn the optimal parameters $\theta^*$ that minimize a predefined loss function $\ell$, e.g., cross-entropy or mean squared error. Formally, we have:
\begin{equation}  
\theta^* = \arg\min_{\theta} \frac{1}{N} \sum_{i=1}^{N} \ell(f_\theta(x_i), y_i).
\label{eq:ERM}
\end{equation}

\myhead{Data Error Types} 
In this paper, we consider the following three error types that are prevalent in real-world training data.

    (1) \textbf{Label Error (LE).} 
    Let $y_i^*$ denote the ground-truth label for training sample $z_i$. We define the set of samples with label errors as $\mathcal{D}_{t}^{\rm LE}=\{ z_i=(x_i, y_i) \mid y_i \neq y_i^* \}$. Label errors are ubiquitous and typically arise from imperfect annotation processes, including human subjectivity in crowdsourced tasks and faults in automated labeling functions~\cite{Trustworthy_AI_Data_NatMachIntell22}.

    (2) \textbf{Feature Error (FE).} 
    Let $x_i^*$ denote the ground-truth feature of sample $z_i$. The set of samples with feature errors is denoted as $\mathcal{D}_{t}^{\rm FE}=\{ z_i=(x_i, y_i) \mid x_i \neq x_i^* \}$. Feature errors manifest in various forms, such as missing values, outliers, and attribute dependency violations. These anomalies typically arise from sporadic or systematic failures during data collection~\cite{UniClean_VLDB25}.    

    (3) \textbf{Spurious Correlation (SC).} We first define a group annotation $g:=(y,a)$ composed of a label $y\in\mathcal{Y}$ and a non-causal spurious attribute $a\in\mathcal{A}$~\cite{Spurious_Correlation_KDD25}. 
    The training data can be partitioned into groups based on $g$. When certain groups have significantly smaller sample sizes, we identify them as underrepresented minority groups, denoted as $\mathcal{G}_{\rm min}$. We focus on samples within these underrepresented groups, defined as $\mathcal{D}_{t}^{\rm SC}=\{ z_i=(x_i, y_i) \mid (y_i,a_i) \in \mathcal{G}_{\rm min} \}$. Such an imbalance typically arises from sampling bias, causing models to learn spurious attributes as non-causal shortcuts~\cite{Spurious_Correlation_KDD25}. 

\begin{figure}[t]
    \centering
    \includegraphics[width=1\linewidth]{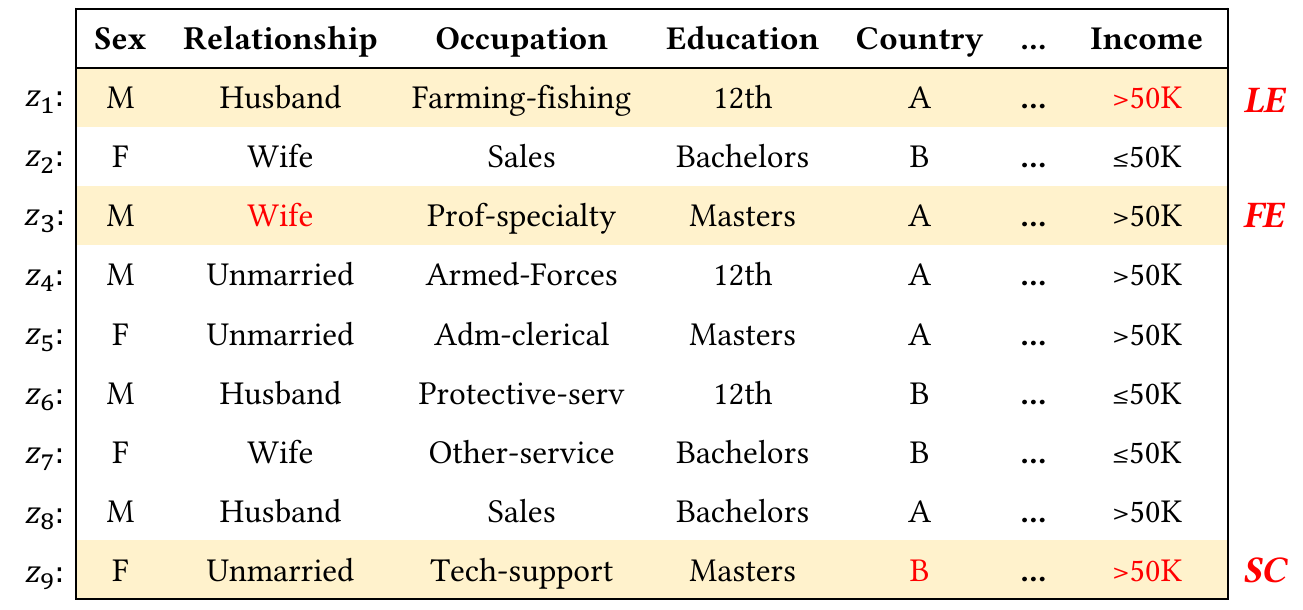}
    \caption{An example of three error types\protect\footnotemark{} (Adult dataset).}
    \label{fig:example}
\end{figure}

\begin{example}
  Fig.~\ref{fig:example} illustrates three data error types in the Adult dataset for the income prediction task.
  (1) LEs stem from incorrect annotation: for instance, $z_1$ with the golden label ``$\leq$50K'' is mislabeled as ``$>$50K'';
  (2) $z_3$ exhibits FE for violating the attribute dependency between ``Sex'' and ``Relationship'';
  (3) Regarding SCs, with ``Country'' as a spurious attribute, majority groups (``Country=A, Income$>$50K'' and ``Country=B, Income$\leq$50K'') establish a non-causal spurious correlation between ``Country'' and ``Income". ML models learned through ERM typically exploit this shortcut, resulting in prediction failures on underrepresented samples like $z_9$ (with ``Country=B, Income$>$50K'').
\end{example}

\myhead{Data with Mixed Error Types}
    We consider training data $\mathcal{D}_{t}$ contaminated by a mixture of error types $\mathcal{T}=\{\rm LE, FE, SC \}$. Formally, the set of erroneous samples is defined as the union of individual error sets: $\mathcal{D}^{\rm Err}_{t} = \mathcal{D}^{\rm LE}_{t} \cup \mathcal{D}^{\rm FE}_{t} \cup \mathcal{D}^{\rm SC}_{t}$. This setting explicitly allows for \textit{error co-occurrence}, meaning any single sample may simultaneously contain multiple types of errors. 

\begin{definition}[Training data debugging]
    Let $\mathcal{T}=\{\text{LE, FE, SC}\}$ be the set of potential error types. \emph{Training data debugging} aims to learn a mapping $g: \mathcal{X}\times \mathcal{Y} \to \{0,1\}^{3}$. For each sample $z_i$, $g$ assigns a binary error vector $\hat{{\rm t}}_i = [\hat{t}_{i}^{(1)},\dots, \hat{t}_{i}^{(3)}]$, where $\hat{\rm t}_{i}^{(k)}=1$ indicates the presence of the $k$-th error type in $\mathcal{T}$. Clearly, the mapping function indicates both erroneous samples (where $\hat{{\rm t}}_i \neq \mathbf{0}$) and their error types.

\end{definition}

\myhead{Influence Function}
\label{section:influence_function}
To quantify the impact of a training sample $z_i$ on a validation sample $z_j$, the {Leave-One-Out (LOO)} score offers a straightforward influence by computing: $\text{LOO}(z_i, z_j) := \ell(z_j; \hat{\theta}_{-i})-\ell(z_j; \hat{\theta})$. However, computing exact LOO scores is computationally prohibitive as it requires retraining the model for each sample.
Consequently, Influence Functions~\cite{influence_ICML17, Influence_Analysis_Survey_24, Grass_influence_NIPS25} are widely adopted as a feasible approximation. By estimating the validation loss change under an infinitesimal up-weighting of $z_i$, the influence score is derived as:
\begin{equation}
\phi(z_i, z_j) := -\nabla_{\hat{\theta}} \ell(z_j; \hat{\theta})^\top H_{\hat{\theta}}^{-1} \nabla_{\hat{\theta}} \ell(z_i; \hat{\theta}),
\label{eq:influence_function}
\end{equation}
where $H_{\hat{\theta}}$ is the empirical Hessian. Intuitively, this approximates the parameter update via a Newton step and estimates the loss change using a first-order Taylor expansion~\cite{Influence_GPT_NIPS25}.
Nevertheless, inverse Hessian remains computationally expensive, particularly for models with massive parameter spaces, e.g., LLMs. To ensure scalability, a common practice is to restrict the influence computation to the parameters of the final LLMs classification layer~\cite{Data_Attribution_Survey25, LESS_ICML24, Grass_influence_NIPS25}.

\footnotetext{Note that SC is a dataset-level error, so we flag minority-group samples (e.g., $z_9$) for the recognition of SC and group-aware reweighting during repair.}

\section{Methodology}
In this section, we present an automated training data debugging framework named \methodname. We begin by analyzing why influence vectors can effectively distinguish training samples with different error types. Building on this insight, we develop a parameterized classifier with a specialized training paradigm. We then describe how these components integrate into a complete debugging pipeline.

\subsection{Influence Vector}

Different data error types affect model training in distinct mechanisms. Specifically, label errors corrupt supervision signals and distort decision boundaries; feature errors perturb input representations and the feature space geometry; spurious correlations encourage models to rely on non-causal shortcuts. 
Our key observation is that these mechanistic differences produce distinct behavior patterns on the validation set through influence functions. A training sample with a specific error type tends to influence particular subsets of validation samples in a characteristic way. For example, it may predominantly affect samples with similar features, samples near decision boundaries, or samples within the same spurious subgroup. These structured patterns provide direct signals for distinguishing error types.

To capture such patterns, we represent each training sample by its influence vector, which records its influence on every validation sample. 
Formally, for a training sample $z_i$, we define its influence vector $\Phi_i \in \mathbb{R}^M$ as the concatenation of its influence functions over the validation set $\mathcal{D}_{v}$:
\begin{equation}
\Phi_i := [\phi_{i,1}, \dots, \phi_{i,M}],
\label{eq:influence_vector}
\end{equation}
where $M = |\mathcal{D}_{v}|$ and each entry $\phi_{i,j}=\phi(z_i, z_j)$ is the influence function of $z_i$ w.r.t. a validation sample $z_j$.

Prior data attribution methods~\cite{influence_ICML17, Influence_Relabeling_ICLR21, Data_Glitches_KDD25, LESS_ICML24} typically summarize influence as a single scalar value, collapsing structured patterns and obscuring error-specific signals. In contrast, the complete influence vector preserves both the magnitude and distribution of influence, making it effective for identifying different error types.


\begin{example}
\label{example}
    We illustrate the effectiveness of influence vectors using the examples in Fig.~\ref{fig:example}.
   (1) LE: Since $z_1$ violates feature-label mapping, the model tends to memorize this anomaly. Consequently, it misguides predictions on validation samples $z_j$ that share similar features but hold the correct label. Removing $z_1$ would significantly reduce the loss on $z_j$, manifesting as a strongly negative influence value, whereas its impact on dissimilar samples remains negligible.
   (2) FE: Since $z_3$ violates the dependency between ``Sex'' and ``Relationship'', the model treats it as an outlier isolated from class centers. Such outliers exert a minor effect on the decision boundary. Consequently, in the influence vector, $z_3$ exhibits fluctuating influence on validation samples very close to the boundary, while having a negligible impact on samples well within the class center.
   (3) SC: Due to spurious correlations, the model may exploit ``Country'' as a shortcut to predict the ``Income''. In terms of influence, the underrepresented sample $z_9$ exerts a significant impact on validation samples within the same group, whereas its impact on samples from other groups is negligible.
   \camera{We further present a case study on the Adult dataset in Appendix~\ref{app: case}, qualitatively showing that influence vectors exhibit distinct patterns for different error types.}
\end{example}

\subsection{Data Debugging via Influence Vectors}
We reformulate training data debugging as a multi-label classification task with influence vectors as inputs. Formally, given the influence vector $\Phi_i\in\mathbb{R}^M$ and the set fo error types $\mathcal{T}=\{\rm LE, FE, SC\}$, we aim to learn a mapping function $g_{\psi}: \mathbb{R}^M \to \{0, 1\}^{3}$. 
For each sample $z_i\in \mathcal{D}_t $, $g_{\psi}$ outputs a binary vector $\hat{{\rm t}}_i \in \{0, 1\}^{3}$, with each entry indicating the presence of the corresponding error type in $\mathcal{T}$. This formulation enables the detection of multiple error types within a single sample $z_i$, i.e., $\sum_{k=1}^3 \hat{\rm t}_{i}^{(k)} \ge 2$.

We then employ a Data Error Classifier (DEC) for this task.
The architectural design of DEC is driven by two requirements:
(1) \textit{Permutation Invariance}: the influence vector essentially represents an unordered set of scalar values, where the permutation of validation samples holds no semantic information, and (2) \textit{Element-wise Interactions}: The elements within an influence vector are not independent. For instance, a training sample has similar influence values w.r.t. similar validation samples. Capturing such element-wise interactions is crucial for distinguishing specific error types.
Hence, we employ the Set Transformer~\cite{Set_transformer} as the encoder of DEC. Specifically, DEC maps the input set $\Phi_i$ to a low-dimensional representation ${\rm h}_i$ using a unified architecture that stacks a Set Attention Block (SAB) followed by Pooling by Multihead Attention (PMA) with $l$ learnable seed vectors $S$:
\begin{equation}
    {\rm h}_i = \text{PMA}_l(\text{SAB}(\Phi_i)) = \text{MAB}(S, \text{rFF}(\text{MAB}(\Phi_i, \Phi_i))).
    \label{eq:set_transformer}
\end{equation}
The MLP-based heads parameterized by $\omega$ then map ${\rm h}_i$ into predictions: $\hat{\mathbf{t}}_i = \text{MLP}_{\omega}({\rm h}_i)$.

The core building block Multihead Attention Block (MAB) processes two input sets $X$ and $Y$ through:
\begin{equation}
    \text{MAB}(X, Y) = \text{LayerNorm}(H + \text{rFF}(H)),
\end{equation}
where $H = \text{LayerNorm}(X + \text{Multihead}(X, Y, Y))$ and $\text{rFF}$ is a row-wise feedforward layer.
Functionally, $\text{SAB}$ preserves permutation equivariance to capture element-wise interactions, while $\text{PMA}$ enforces permutation invariance by aggregating these equivariant features via attention against the fixed seed queries $S$.

\begin{figure}
    \centering
    \includegraphics[width=1\linewidth]{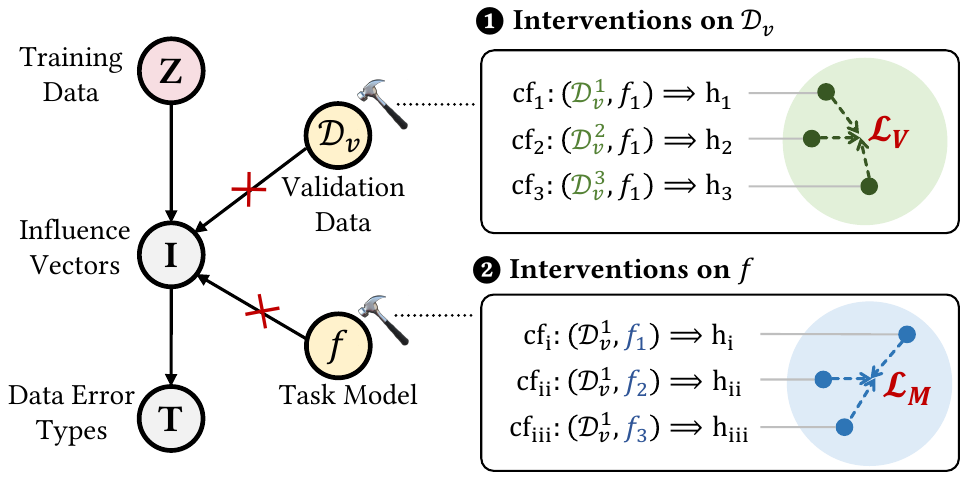}
    \caption{Interventions on the validation set and task model.}
    \label{fig:interventions}
\end{figure}

\subsection{Training Paradigm}
We formalize the training data for DEC as paired instances $\mathcal{E}=\{e_i = (\Phi_i, t_i)\}|_{i=1}^N$.
To construct $\mathcal{E}$, we employ a \textbf{controlled error injection} strategy. Specifically, given an ML task with training set $\mathcal{D}_{t}$, validation set $\mathcal{D}_{v}$, and a task model $f$, we introduce random mixed-type errors into a clean data subset via the following protocols:
(1) LEs are generated by perturbing raw annotations, such as flipping labels for classification or adding noise for regression.
(2) FEs are introduced by compromising feature integrity or violating attribute dependencies, e.g., injecting random noise or disrupting logical constraints.
(3) SCs are synthesized by manipulating subgroup ratios to induce strong correlations between non-causal attributes and target labels.
Crucially, this process explicitly supports {error overlap}, enabling the generation of samples exhibiting compound error types. 
Furthermore, by varying injection ratios and random seeds, we can explicitly control the error intensity and synthesize highly diverse training datasets for DEC tailored to specific tasks.

To optimize DEC for data debugging accuracy, we minimize the standard element-wise Binary Cross-Entropy (BCE) loss. 
Given model predictions $\hat{\rm t}_i$ and labels ${\rm t}_i$, the prediction loss is defined as:
\begin{equation}
    \mathcal{L}_{\rm pred} = - \sum_{k=1}^3 \left[ {\rm t}_{i}^{(k)} \log(\hat{\rm t}_{i}^{(k)}) + (1 - {\rm t}_{i}^{(k)}) \log(1 - \hat{\rm t}_{i}^{(k)}) \right].
    \label{eq:pred_loss}
\end{equation}

However, relying solely on Equation~\eqref{eq:pred_loss} presents a critical challenge. 
The calculation of influence vectors is inherently context-dependent, relying not only on the training sample itself but also on the configurations including the choice of validation set $\mathcal{D}_v$ and the task model $f$. Consequently, standard DEC training risks inadvertently encoding these configuration-relevant artifacts alongside causal patterns related to error types into the learned representations. This causes the DEC to overfit to specific task configurations, thereby impeding generalization. To address this, drawing inspiration from invariant learning~\cite{DIR}, we propose an intervention-based training strategy. As illustrated in Figure~\ref{fig:interventions}, we systematically perturb the validation set selection and model architecture for the same training data, effectively simulating diverse environments. By introducing specific regularization terms, we compel the Set Transformer to learn invariant representations that are robust to different configurations.

(1) \textbf{Interventions on Validation Set.} 
Since influence calculation relies on the validation set, using a fixed set inevitably introduces specific biases unrelated to error types.
To mitigate this, we operate interventions by randomly sampling multiple validation subsets. Specifically, for a training sample $e_i$, we generate augmented views $\{ e_i^{(1)}, \dots, e_i^{(V)} \}$ derived from $V$ distinct validation subsets.
Since the true error type is invariant to the choice of validation set, we enforce representation invariance using an InfoNCE-inspired contrastive loss~\cite{InfoNCE}. Let ${\rm h}_i$ be the anchor embedding and $\{ {\rm h}_i^{(1)}, \dots, {\rm h}_i^{(V)} \}$ be its positive pairs from different validation set. The loss is defined as:
\begin{equation}
    \mathcal{L}_{V} = - \mathbb{E}_{{\rm h}_i, {\rm h}_i^{(v)}} \left[\log \frac{e^{\text{sim}({\rm h}_i, {\rm h}_i^{(v)}) / \tau}}{ e^{\text{sim}({\rm h}_i, {\rm h}_i^{(v)}) / \tau} + \sum_{j \in \mathcal{B}, {\rm t}_j \neq {\rm t}_i} e^{\text{sim}({\rm h}_i, {\rm h}_j) / \tau}} \right],
\end{equation}
where $\text{sim}(\cdot, \cdot)$ denotes cosine similarity, $\tau$ is the temperature hyperparameter, and $\mathcal{B}$ is the current batch. Minimizing $\mathcal{L}_{V}$ effectively pulls the representations of positive pairs closer while pushing negative pairs apart. This optimization compels DEC to distill a representation ${\rm h}_i$ that is invariant to the validation set and solely dependent on patterns related to data error labels.

(2) \textbf{Interventions on Task Model.} 
Similarly, the task model architecture constitutes another configuration for influence vectors. Different architectures produce influence vectors with vastly different statistical distributions (e.g., scale and sparsity), creating distribution shifts that hinder transferability. 
To ensure DEC learns representations invariant to task model architecture evolving, we simulate diverse environments by constructing an ensemble of $K$ heterogeneous models, $\{f_1, \dots, f_K\}$. Consequently, for each training sample $z_i$, we compute a set of $K$ influence vectors $\{\Phi_i^{(1)}, \dots, \Phi_i^{(K)}\}$, corresponding to the $K$ distinct task models.
To explicitly align the representation spaces across these diverse architectures, we employ an alignment loss. 
Let ${\rm h}_i^{(k)}$ denote the representation derived from the influence vector of the $i$-th sample under the $k$-th model $f_k$. Since all $K$ representations correspond to the same error type, they should be mapped to an identical point in the representation space, regardless of the task model. We therefore minimize the pairwise Euclidean distance between representations of the same sample across all task models. The alignment loss is defined as:
\begin{equation}
    \mathcal{L}_{M} =  \frac{1}{K(K-1)} \sum_{1 \le a \neq b \le K} \left\| {\rm h}_i^{(a)} - {\rm h}_i^{(b)} \right\|^2_2.
\end{equation}
Minimizing $\mathcal{L}_{M}$ penalizes the discrepancy between representations from different task models, compelling the DEC to extract consistent, model-invariant features that are generalizable to task model variations.

The overall objective function of DEC integrates the primary prediction loss with our proposed regularization terms for invariant learning. Formally, the overall loss is defined as:
\begin{equation}
    \label{eq:overall loss}
    \mathcal{L}_{\rm DEC} =  \mathcal{L}_{\rm pred} + \lambda_1 \mathcal{L}_V + \lambda_2 \mathcal{L}_M,
\end{equation}
where $\lambda_1$ and $\lambda_2$ are trade-off hyperparameters that balance the contribution of validation set invariance and task model invariance. We detail the hyperparameter settings in the implementation details of Section~\ref{section:implementation}.

\begin{algorithm}[t]
\caption{Training Data Debugging Workflow in \methodname}
\label{alg:demix}
\begin{algorithmic}[1]
\REQUIRE Training set $\mathcal{D}_{t}=\{ z_i \}_{i=1}^N$, Validation set $\mathcal{D}_{v}=\{ z_j \}_{j=1}^M$, DEC $g_\psi$, repair tools $\mathcal{R}=\{R_t\}_{t \in \mathcal{T}}$
\ENSURE Corrected training set $\mathcal{D}_{t}^{\rm r}$

\STATE Train task model $\hat{\theta}$ on $\mathcal{D}_{t}$
\STATE Initialize $\mathcal{D}_{t}^{\rm r} \leftarrow \emptyset$
\FOR{each training sample $z_i \in \mathcal{D}_{t}$}
    \STATE Compute influence vector $\Phi_i$ on $\mathcal{D}_{v}$ and $\hat{\theta}$ with Equation~\eqref{eq:influence_vector}
    \STATE Predict data error types via DEC: $\hat{\rm t}_i \leftarrow g_\psi(\Phi_i)$

    \IF{$\hat{\rm t}_i = (0,0,0)$}
        \STATE $\mathcal{D}_{t}^{\rm r} \leftarrow \mathcal{D}_{t}^{\rm r} \cup \{z_i\}$ {{\COMMENT{Keep clean samples}}}
    \ELSE
        \STATE $\hat{z}_i \leftarrow z_i$
        \FOR{each $t \in \mathcal{T}$ where $\hat{\rm t}_i$ is activated for $t$}
            \STATE $\hat{z}_i \leftarrow R_t(\hat{z}_i)$ {{\COMMENT{Apply type-specific repair}}}
        \ENDFOR
        \STATE $\mathcal{D}_{t}^{\rm r} \leftarrow \mathcal{D}_{t}^{\rm r} \cup \{ \hat{z}_i \}$
    \ENDIF
\ENDFOR
\RETURN $\mathcal{D}_{t}^{\rm r}$
\end{algorithmic}
\end{algorithm}

\subsection{Overall Debugging Workflow}
The overall debugging workflow of \methodname~is outlined in Algorithm~\ref{alg:demix}. First, \methodname~computes influence vectors for training samples. The influence vectors are then fed into a trained DEC, which identifies erroneous samples and diagnoses their specific error types. Subsequently, \methodname~executes type-specific repair on the detected samples; detailed implementation of these repair tools is provided in the implementation details of Section~\ref{section:implementation}.

\myhead{Complexity Analysis}
Given $N$ training samples, $M$ validation samples, and a task model with $P$ parameters, the time complexity of data debugging comprises three sequential stages: task model training, influence vector computation, and error type classification. First, training the task model for $T$ epochs incurs a complexity of $\mathcal{O}(N \cdot P \cdot T)$. Second, influence vector computation involves estimating the Hessian-vector product for each validation sample via a stochastic algorithm~\cite{influence_ICML17} and multiplying it by the training sample gradients, leading to $\mathcal{O}(N \cdot M \cdot P)$. Finally, the error classifier inference requires approximately $\mathcal{O}(N \cdot M^2)$, dominated by the self-attention mechanism. Consequently, the total time complexity is approximately 
\camera{$\mathcal{O}(N(PT + PM + M^2))$}.
Regarding space complexity, our method requires $\mathcal{O}(P)$ for storing model parameters and $\mathcal{O}(N \cdot M)$ for the influence vectors.

\section{Experiments}
In this section, we evaluate \methodname~through comprehensive experiments, aiming to answer the following research questions:

\noindent \textbf{RQ1 (Debugging performance):} How accurately can \methodname~distinguish erroneous data along with their error types, particularly under different ratios of data errors?

\noindent\textbf{RQ2 (Repair efficacy):} Can \methodname~be effectively utilized to repair data and improve task model performance?

\noindent\textbf{RQ3 (Ablation studies):} What is the specific contribution of each component within \methodname~to its overall effectiveness?

\subsection{Experimental Setup}
\label{section:exp}
\myhead{Datasets and Models} 
\camera{To the best of our knowledge, no existing real-world dataset simultaneously contains mixed error types with known error-type labels, we therefore adopt controlled error injection into clean base data for systematically evaluation of data debugging methods. To demonstrate the versatility of the evaluation, we select clean base datasets spanning three diverse domains:}
(1) Tabular data prediction: We select six standard benchmark datasets from the UCI repository\footnote{https://archive.ics.uci.edu/dataset/}, covering diverse task types: binary classification (\textbf{Adult}, \textbf{Bank}, and \textbf{Credit}), multi-class classification (\textbf{Covertype}), and regression (\textbf{Bike Sharing} and \textbf{Air Quality}). For task model architectures, we employ 3 models: a 2-layer MLP1, a 4-layer MLP2 and a state-of-the-art Transformer-based model, TabPFN~\cite{TabPFN}.
(2) Recommendation systems: We utilize three widely adopted datasets: \textbf{Amazon}\footnote{https://nijianmo.github.io/amazon/index.html}, \textbf{MovieLens}\footnote{https://grouplens.org/datasets/movielens/}, and \textbf{Yelp}\footnote{https://www.kaggle.com/datasets/yelp-dataset/yelp-dataset}. For task model architectures, we employ three deep recommendation models DIN~\cite{DIN}, DIEN~\cite{DIEN} and DHEN~\cite{DHEN}.
(3) LLM Alignment: To evaluate the scalability of \methodname~to large-scale generative models, we extend our evaluation to the LLM preference alignment task. We employ the \textbf{UltraFeedback}\footnote{https://huggingface.co/datasets/trl-lib/ultrafeedback\_binarized} and \textbf{Capybara}\footnote{https://huggingface.co/datasets/trl-lib/Capybara-Preferences} preference datasets to fine-tune a Qwen-0.5B-Instruct~\cite{qwen2.5} model via Direct Preference Optimization (DPO)~\cite{DPO}. The alignment quality of the task model is subsequently evaluated on the AlpacaEval benchmark\footnote{https://github.com/tatsu-lab/alpaca\_eval}.

\camera{Beyond synthetic injected data errors, to verify the practical validity of data debugging methods on datasets with naturally data errors, we introduce two publicly available datasets: \textbf{CIFAR-10N}~\footnote{https://github.com/UCSC-REAL/cifar-10-100n} provides human-annotated LEs for CIFAR-10 images, and \textbf{CelebA}~\footnote{https://www.kaggle.com/datasets/jessicali9530/celeba-dataset} is a face attribute prediction dataset with natural SC between gender and hair color.}

\myhead{Controlled Error Injection}
\label{section:error_injection}
To verify the effectiveness of the data debugging method, we inject data errors into the datasets and expect the debugging methods to accurately identify and repair these injected errors. 
We first split all datasets into training, validation, and test sets with a ratio of $(0.6, 0.2, 0.2)$, and then inject random mixed-type errors into the training and validation sets. \textit{We retain a proportion $\alpha$ of raw data as clean samples}. For the remaining $(1-\alpha)$, we introduce various error types, allowing error overlap, where a single sample may simultaneously harbor multiple error types. Injection of specific error types is conducted as follows:

(1) {Label Errors (LEs)}: For tabular data classification tasks, we flip the ground-truth label to a different class; For tabular data regression tasks, we add random noise to the original labels. For recommendation tasks, where labels are typically binary (indicating user-item interactions such as clicks or reviews), we flip the interaction status (e.g., $1 \to 0$). For LLM alignment tasks, which rely on preference pairs (``chosen'' vs. ``rejected''), we inject errors by swapping the ``chosen'' and ``rejected'' responses along with their corresponding reward scores.

(2) {Feature Errors (FEs)}: For tabular data prediction and recommendation systems, where data typically consists of various feature columns, we randomly select several feature columns and inject different types of feature errors, including outliers, attribute dependency violations, categorical feature flipping, and format inconsistencies~\cite{UniClean_VLDB25}. Additionally, for sequential features in recommendation (e.g., user interaction history), we apply random shuffling, deletion, or replacement. For LLM alignment, we introduce feature errors by randomly replacing or deleting tokens in the response text.

(3) {Spurious Correlations (SCs)}: To induce reliance on non-causal shortcuts, we first construct a set of spurious attributes $\mathcal{A}$ following~\cite{Spurious_tabular_ICML25}. 
We randomly select one spurious attribute $a \in \mathcal{A}$ and ensure that the proportions of majority and minority groups with respect to $a$ are approximately 90\% and 10\%, respectively. The choice of attributes varies by domain: for tabular and recommendation data, we leverage non-causal attributes such as gender or race~\cite{Spurious_tabular_ICML25}; for LLM preference data, we use attributes that are not directly related to response quality, such as response length, format structure, emoji, etc.~\cite{chen2024odin, From_list_to_emoji_ACL25}. 


\myhead{Repair Tools}
\label{app:repair}
To repair the identified errors, we employ specific repair strategies tailored to each error type. 
(1) For LEs, the repair mechanism depends on the task: regarding tabular data classification and recommendation systems, following~\cite{Influence_Relabeling_ICLR21, Cleanlab_label}, we correct noisy labels by replacing them with the class indices maximizing the predicted probabilities derived from $k$-fold cross-validation; regarding regression tasks, we essentially discard the detected LE samples to prevent skewing the decision boundary; regarding LLM alignment tasks, we switch the ``chosen'' and ``reject'' responses.
(2) For FEs, following~\cite{UniClean_VLDB25, Cleanlab_ood} we adopt a statistical imputation approach based on Z-scores for tabular data prediction and recommendation systems. We first compute the mean and standard deviation of features using only clean samples (per-class for classification, global for regression). Feature values in erroneous samples with Z-scores exceeding a threshold are deemed outliers and replaced by the corresponding mean value. Regarding LLM alignment, we discard samples with FEs.
(3) Finally, for SCs, rather than modifying feature values, we implement a sample reweighting strategy following~\cite{Spurious_Correlation_KDD25}. To mitigate the model's reliance on majority shortcuts, we assign a higher importance weight to samples from minority groups, thereby forcing the model to focus more on these underrepresented instances during training.

\myhead{Evaluation Metrics}
We evaluate the effectiveness of data debugging methods from two key dimensions:(1) Debugging F1: We employ the average F1 scores to assess the method's precision in identifying specific error types within mixed-error data. For each specific error type, we first calculate the F1-score $F1_t$ for each specific error type $t \in \mathcal{T}$, and the overall debugging performance is quantified by the average of $F1_t$ across all error types;
(2) Repair efficacy: We report the task model's performance (Accuracy and MSE for tabular data prediction; AUC for recommendation; and WinRate against GPT4 for LLM alignment) on the test set after training on the repaired training data. 

\myhead{Baselines}
(1) \textbf{Clean Data:} The unperturbed raw data;
(2) \textbf{Error. Data:} The data with mixed data error types through controlled error injection;
(3) \textbf{Debugging with Data Cleaning (DDC):} 
Cleanlab~\cite{Cleanlab_label, Cleanlab_ood} provides state-of-the-art detection tools for label issues, outliers, and underperforming groups\footnote{https://docs.cleanlab.ai/stable/cleanlab/datalab/guide/issue\_type\_description.html}. Following~\cite{SAGA_SIGMOD23}, we sequentially organize these cleaning tools into a pipeline DDC for debugging data with mixed error types;
(4) \textbf{Debugging with Data Attribution (DDA):} Following state-of-the-art data attribution methods~\cite{Opendataval_NIPS23, Influence_Relabeling_ICLR21}, we construct two baselines: \textbf{DDA-select}, which computes scalar influence scores for training samples by aggregating influence function values across validation data and discards those with low scores~\cite{Opendataval_NIPS23, LESS_ICML24}; and \textbf{DDA-repair}, which first identifies detrimental samples with influence scores, then clusters their influence vectors to infer error types based on heuristics~\cite{Influence_Relabeling_ICLR21, Data_Glitches_KDD25, Shapley_negative_clustering_ICML24, Influence_embedding_NIPS23}, enabling type-specific repair. Note that debugging F1 can only be computed for DDA-repair, as DDA-select performs binary selection without debugging.
(5) \textbf{\methodname}: Our framework uses a DEC for error type classification, followed by type-specific repair. In \methodname, we train a specific DEC for each dataset and use it for debugging. We further propose \textbf{\methodname-unif}, a variant where a unified DEC is trained with a diverse dataset collection and directly deployed to target datasets for debugging. 
More details about the workflow of baselines are provided in Appendix~\ref{app:workflow}.

\myhead{Implementation Details} 
\label{section:implementation}
(1) We implement the DEC using a Set Transformer with 2 SAB layers and 1 PMA layer, followed by 3 MLP-based task heads. 
For \methodname, we set $|\mathcal{E}| = 50,000$ per dataset; for \methodname-unif, the unified DEC is trained on a composite dataset with $|\mathcal{E}| = 20,000$ samples drawn from each source dataset.
For interventions on the validation set, we set $V=10$; for interventions on the task model, we set $K=3$ for tabular data prediction and recommendation systems, $K=1$ for LLM alignment.
Regarding loss coefficients, we set $\lambda_1=0.1$ and $\lambda_2=0.1$.
(2) As for data repair tools, we apply targeted strategies for each type of data errors: LEs are repaired via probability-based relabeling~\cite{Influence_Relabeling_ICLR21}; FEs are repaired by locating and correcting perturbed feature columns~\cite{UniClean_VLDB25}; and SCs are mitigated by up-weighting identified minority samples during retraining~\cite{Spurious_Correlation_KDD25, JTT_ICML21}. Details about repair tools are available in Appendix~\ref{app:repair}.
We keep the repair strategies the same for different debugging baselines. 
(3) All experiments were conducted on a server equipped with a Montage Jintide\textsuperscript{\textregistered} C6226R CPU, 256GB RAM, and four NVIDIA GeForce RTX 4090 GPUs. 

\begin{figure}[t]
    \centering
    \includegraphics[width=1\linewidth]{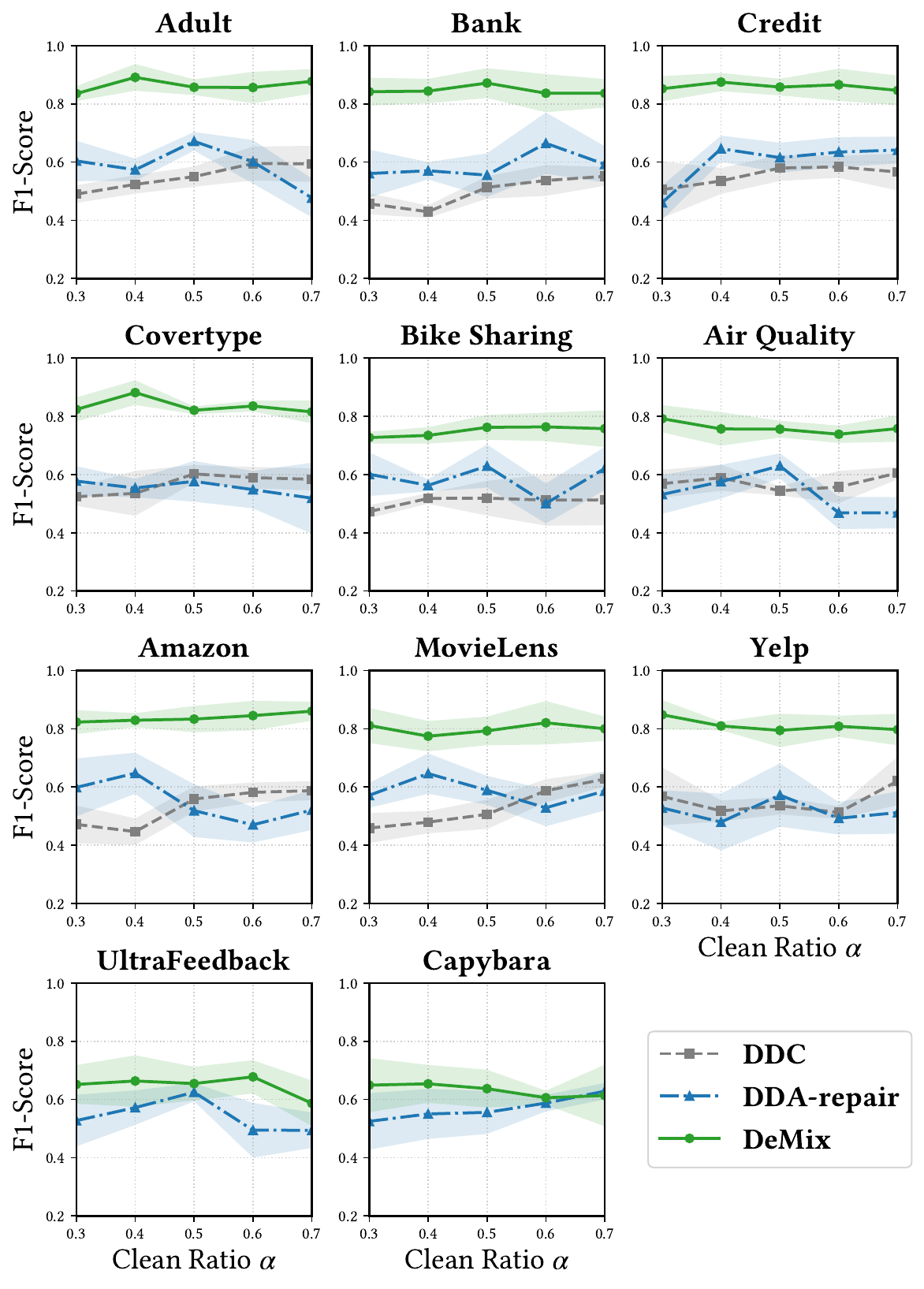}
    \caption{Debugging F1-score (\%) on 11 datasets across 5 independent runs under varying clean data ratios $\alpha$. }
    \label{fig:diagnosis_f1}
\end{figure}

\begin{figure}[t]
    \centering
    \includegraphics[width=0.95\linewidth]{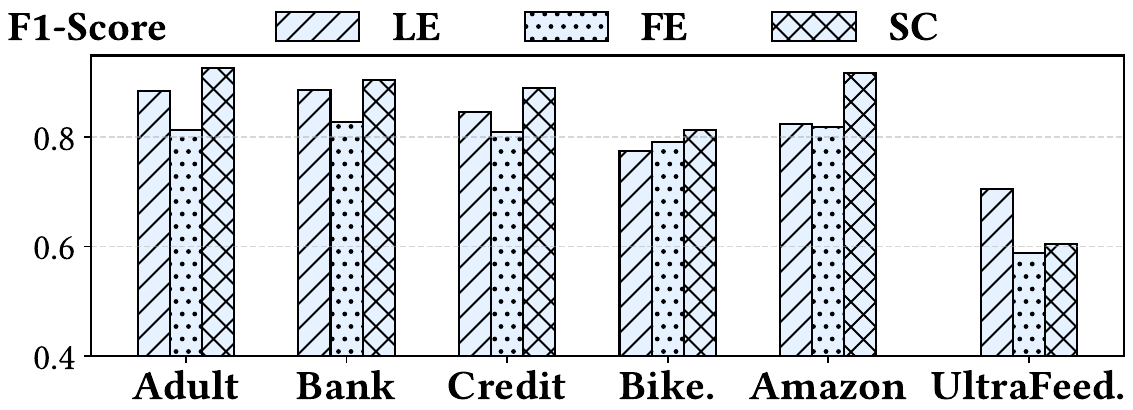}
    \caption{Debugging F1-score (\%, $\alpha=0.5$) of \methodname~for specific error types.}
    \label{fig:diagnosis_bar}
\end{figure}

\subsection{Debugging Performance}
To answer \textbf{RQ1}, we first analyze the average debugging F1 across all error types and for each specific error type, followed by a granular analysis on hard cases when a single training sample contains multiple error types.

\myhead{Overall Performance} 
\label{section:exp_f1}
As illustrated in Fig.~\ref{fig:diagnosis_f1}, \methodname~shows superior debugging F1 compared to state-of-the-art baselines. We summarize the key observations as follows:
(1) \textit{Effectiveness of \methodname:} \methodname~consistently outperforms existing baselines across nearly all datasets and $\alpha$. Quantitatively, it achieves an average improvement of \textbf{22.61\%} in F1-score. This substantial margin validates the capability of \methodname~in accurately data debugging, irrespective of data formats (tabular, recommendation, or text) or error ratios.
(2) \textit{Robustness of \methodname:} A distinguishing feature of \methodname~is its stability against variations in error ratios. As observed in Fig.~\ref{fig:diagnosis_f1}, while baseline performance fluctuates significantly with changes in $\alpha$ (e.g., DDA-repair on {Adult} and {Bank}), \methodname~maintains a consistently high performance curve with minimal variance. This stability suggests that \methodname~successfully captures the intrinsic patterns of data errors rather than overfitting to distributional statistics that shift with error ratios, thereby ensuring reliable data debugging in dynamic real-world scenarios.
(3) \textit{Type-specific performance}:
Figure~\ref{fig:diagnosis_bar} details the debugging performance of \methodname~across specific error types. Overall, \methodname~excels at detecting SCs and LEs but shows lower sensitivity to FEs, as they are inherently stealthy and induce weaker influence signals. This limitation is notably observed in LLM alignment tasks, where debugging F1 gains are modest. We believe this results from the approximation error of influence function estimation in large-scale models such as Llama.

\begin{table}[t]
  \centering
  \caption{Debugging F1-score (\%, $\alpha=0.5$) on hard cases where a single sample contains multiple error types.}
  \label{tab:hard_samples}
  \resizebox{\linewidth}{!}{
  \begin{tabular}{lcccccc}
    \toprule
    \textbf{Method} & \textbf{Adult} & \textbf{Bank} & \textbf{Credit} & \textbf{Bike.} & \textbf{Amazon} & \textbf{UltraFeedback} \\
    \midrule
    DDC & 35.37 & 27.63 & 33.15 & \textbf{39.33} & 39.79 & - \\
    DDA-repair & 36.39 & 32.01 & 39.86 & 35.45 & 45.72 & 16.09 \\
    \methodname & \textbf{47.85} & \textbf{42.07} & \textbf{43.50} & 37.02 & \textbf{48.02} & \textbf{19.03} \\
    \bottomrule
  \end{tabular}
  }
\end{table}

\myhead{Analysis on Hard Cases}
To further evaluate the robustness of \methodname, we analyze the performance on a subset of hard cases where a single sample containing multiple co-occurring error types, i.e., for a sample $z_i$, $\sum_{k=1}^3 \hat{\rm t}_{i}^{(k)} \ge 2$. 
As shown in Table~\ref{tab:hard_samples}, \methodname~demonstrates superior capability in handling these hard cases. It achieves the best performance on 5 out of 6 datasets and provides an average improvement of {21.21\%} in debugging F1-score over baselines.
While baselines like DDA-repair rely on scalar influence scores that often produce conflicting signals when errors overlap, \methodname~treats debugging as a multi-label classification task. By leveraging the high-dimensional semantic information within influence vectors, our method can effectively disentangle mixed error patterns.

\begin{table}[t]
  \small
  \centering
  \caption{Debugging F1-score (\%) on datasets without error injection.}
  \label{tab:real_world}
  \begin{tabular}{lcc}
    \toprule
    \textbf{Method} & \textbf{CIFAR-10N (F1 on LE)} & \textbf{CelebA (F1 on SC)} \\
    \midrule
    DDA-repair & 85.31 & 55.97 \\
    \methodname & \textbf{87.95} & \textbf{71.82} \\
    \bottomrule
  \end{tabular}
\end{table}

\myhead{Practical Validity}
\camera{
The practical effectiveness of \methodname~ when deployed on real world applications are two-fold: 
(1)	Detecting naturally occurring errors. As shown in Table~\ref{tab:real_world}, we evaluated \methodname~on datasets with no synthetic error injection. The debugging F1 scores shows that \methodname~consistently outperforms the strongest baseline (DDA-repair) in identifying naturally occurring errors. As a case study in CelebA (with SC between gender and hair color), DeMix achieves high F1 in detecting minority samples (e.g., blond males), highlighting its practical diagnostic capability.
(2) Robustness to noisy base data. In practice, the DEC requires only a small clean base data for error injection (e.g., 10K samples on the Amazon dataset yield >83\% debugging F1-score), which can be directly sourced from the clean validation set. Critically, even when the base data itself is noisy (30\% pre-existing random errors), DeMix's debugging F1 drops by only 2.15\%, demonstrating robustness to realistic deployment conditions where perfectly clean base data may not be available.
}

\begin{table*}[t]
    \centering
    \caption{Performance comparison of data repair across 11 datasets from 3 domains with $\alpha=0.5$. We report the mean and standard deviation of performance (\%) over 5 independent runs. The best results are in \textbf{bold}.}
    \label{tab:repair_perf}
    \resizebox{\textwidth}{!}{ 
    \begin{tabular}{l ccc c cc ccc cc}
        \toprule
        \multirow{3}{*}{\textbf{Method}} & \multicolumn{6}{c}{\textbf{Tabular Data Prediction}} & \multicolumn{3}{c}{\textbf{Recommendation}} & \multicolumn{2}{c}{\textbf{LLM Alignment}} \\
        \cmidrule(lr){2-7} \cmidrule(lr){8-10} \cmidrule(lr){11-12}
         & \textbf{Adult} & \textbf{Bank} & \textbf{Credit} & \textbf{Covertype} & \textbf{Bike } & \textbf{Air} & \textbf{Amazon} & \textbf{MovieLens} & \textbf{Yelp} & \textbf{UltraFeedback} & \textbf{Capybara} \\
         & (Acc$\uparrow$) & (Acc$\uparrow$) & (Acc$\uparrow$) & (Acc$\uparrow$) & (MSE$\downarrow$) & (MSE$\downarrow$) & (AUC$\uparrow$) & (AUC$\uparrow$) & (AUC$\uparrow$) & (WinRate$\uparrow$) & (WinRate$\uparrow$) \\
        \midrule
        Clean Data & $85.78_{\pm 0.10}$ & $90.47_{\pm 0.03}$ & $81.35_{\pm 0.04}$ & $79.48_{\pm 0.51}$ & $1.07_{\pm 0.02}$ & $0.09_{\pm 0.01}$ & $79.38_{\pm 0.32}$ & $81.75_{\pm 0.19}$ & $82.68_{\pm 0.23}$ & $27.15_{\pm 4.24}$ & $28.34_{\pm 3.52}$ \\
        Error. Data & $75.29_{\pm 1.63}$ & $81.50_{\pm 2.37}$ & $72.23_{\pm 2.26}$ & $71.91_{\pm 1.37}$ & $1.61_{\pm 0.11}$ & $0.17_{\pm 0.03}$ & $67.17_{\pm 0.75}$ & $70.93_{\pm 0.42}$ & $72.63_{\pm 0.27}$ & $24.38_{\pm 3.46}$ & $25.52_{\pm 2.95}$ \\
        \midrule
        DDC & $78.83_{\pm 0.44}$ & $85.27_{\pm 0.15}$ & $77.15_{\pm 0.13}$ & $75.43_{\pm 0.46}$ & $1.42_{\pm 0.04}$ & $0.12_{\pm 0.03}$ & $73.92_{\pm 0.47}$ & $76.62_{\pm 0.15}$ & $77.25_{\pm 0.19}$ & - & - \\
        DDA-select & $79.27_{\pm 0.13}$ & $85.58_{\pm 0.27}$ & $76.53_{\pm 0.11}$ & $74.12_{\pm 0.34}$ & $1.31_{\pm 0.04}$ & $0.13_{\pm 0.01}$ & $74.16_{\pm 0.24}$ & $75.30_{\pm 0.24}$ & $77.39_{\pm 0.26}$ & $25.37_{\pm 4.13}$ & $25.53_{\pm 2.15}$ \\
        DDA-repair & $78.53_{\pm 0.21}$ & $83.89_{\pm 0.14}$ & $74.84_{\pm 1.37}$ & $75.03_{\pm 0.36}$ & $1.44_{\pm 0.06}$ & $0.15_{\pm 0.02}$ & $74.03_{\pm 0.25}$ & $76.25_{\pm 0.26}$ & $77.44_{\pm 0.27}$ & $25.82_{\pm 1.92}$ & $24.57_{\pm 2.03}$ \\
        \midrule
        \methodname & $\mathbf{84.91_{\pm 0.24}}$ & $\mathbf{90.02_{\pm 0.15}}$ & $\mathbf{80.12_{\pm 1.04}}$ & $\mathbf{78.62_{\pm 0.82}}$ & $\mathbf{1.11_{\pm 0.03}}$ & ${0.12_{\pm 0.02}}$ & ${79.52_{\pm 0.23}}$ & $\mathbf{81.06_{\pm 0.17}}$ & $\mathbf{81.57_{\pm 0.18}}$ & $\mathbf{26.04_{\pm 2.87}}$ & ${27.53_{\pm 3.74}}$ \\
        \ \ w/o. repair & $82.88_{\pm 0.17}$ & $88.62_{\pm 0.06}$ & $79.31_{\pm 0.65}$ & $77.05_{\pm 0.59}$ & $1.19_{\pm 0.02}$ & $0.14_{\pm 0.02}$ & $77.52_{\pm 0.14}$ & $79.04_{\pm 0.13}$ & $79.55_{\pm 0.15}$ & $26.02_{\pm 2.21}$ & $26.51_{\pm 3.59}$ \\
        \midrule
        \methodname-unif & $83.76_{\pm 0.13}$ & $89.24_{\pm 0.14}$ & $79.05_{\pm 1.02}$ & $77.89_{\pm 0.81}$ & $1.16_{\pm 0.04}$ & $\mathbf{0.10_{\pm 0.02}}$ & $\mathbf{79.61_{\pm 0.25}}$ & $80.28_{\pm 0.16}$ & $80.72_{\pm 0.17}$ & $25.84_{\pm 2.62}$ & $\mathbf{27.95_{\pm 3.71}}$ \\
        \ \ w/o. repair & $81.73_{\pm 0.21}$ & $87.97_{\pm 0.18}$ & $78.02_{\pm 0.75}$ & $76.96_{\pm 0.59}$ & $1.22_{\pm 0.03}$ & $0.12_{\pm 0.01}$ & $79.61_{\pm 0.17}$ & $78.25_{\pm 0.26}$ & $79.69_{\pm 0.11}$ & $25.97_{\pm 2.74}$ & $26.82_{\pm 3.65}$ \\
        \bottomrule
    \end{tabular}
    }
\end{table*}

\subsection{Repair Efficacy}
To answer \textbf{RQ2}, we conduct type-specific data repair based on the classification results from DEC and use the repaired data to retrain the task model and observe its test performance.
Table~\ref{tab:repair_perf} presents a comprehensive performance evaluation of \methodname~across 11 datasets covering 3 distinct domains. In these experiments, we fix $\alpha=0.5$. For task models, we adopt MLP1 for tabular prediction, DIN for recommendation, and Qwen2.5-0.5B-Instruct for LLM alignment. The experimental results lead to the following key observations:
(1) \textit{Impact of mixed error types:} 
The introduction of mixed error types significantly degrades the performance of task models. As evidenced by the results of the Error. Data without any data debugging (second row), models trained on such contaminated data exhibit substantial performance degradation (19.60\% on average) compared to the Clean Data baseline. This considerable gap underscores the critical need for effective training data debugging methods in practical ML pipelines.
(2) \textit{Effectiveness of \methodname:} 
Our proposed \methodname~consistently outperforms other data debugging baselines. On average, \methodname~surpasses all data debugging baselines by 9.32\% on average, demonstrating that our framework can perform effective debugging for type-specific repair.
(3) \textit{Transferability of \methodname-unif:} 
While \methodname~generally achieves the highest performance due to its task-specific specialization, \methodname-unif remains highly competitive. Notably, \methodname-unif attains the best performance on the Air Quality and Amazon datasets. This indicates that our DEC is capable of capturing transferable data error patterns across datasets without a significant performance drop.
(4) \textit{Importance of data repair:} 
Results of \methodname~w/o. repair and \methodname-unif w/o. repair (the 7th and 9th rows of Table~\ref{tab:repair_perf}) reveal that simply discarding erroneous samples is suboptimal. The full \methodname~framework, which incorporates data repair, consistently outperforms the removal-only variant. This confirms that data repair is crucial for recovering valuable information from erroneous samples.
(5) \textit{Limitations of baselines:} 
We observe that DDA-repair underperforms DDA-select on most datasets. Correlating this observation with Section~\ref{section:exp_f1}, we attribute this deficiency to the limited debugging F1 of DDA in data with mixed error types. Data repair based on inaccurate debugging results introduces additional noise, thereby undermining model training. Furthermore, DDC provides only marginal improvements, indicating that data cleaning is inadequate for handling data with complex systematic errors.

\subsection{Ablation Studies}
To answer \textbf{RQ3}, we conducted ablation studies to evaluate the marginal contributions of $\mathcal{L}_V$ and $\mathcal{L}_M$.
For tabular data classification tasks, we utilize three task models (MLP1, MLP2, and TabPFN) and two disjoint validation sets, $\mathcal{D}_{v}^1$ and $\mathcal{D}_{v}^2$.
We constructed three configurations for influence vector computation: ${\rm cf}_1$ ($\mathcal{D}_{v}^1$ and MLP1\&MLP2), ${\rm cf}_2$ ($\mathcal{D}_{v}^2$ and MLP1\&MLP2), and ${\rm cf}_3$ ($\mathcal{D}_{v}^1$ and TabPFN). The DEC is trained on influence vectors computed on ${\rm cf}_1$, and evaluated on ${\rm cf}_2$ and ${\rm cf}_3$, to simulate real-world scenarios when validation sets and model architectures evolve.
Table~\ref{tab:ablation_cf} presents the results.
First, removing $\mathcal{L}_V$ causes marked performance degradation on ${\rm cf}_2$, with F1 scores dropping by an average of 3.5\% across datasets. This highlights the necessity of $\mathcal{L}_V$ in preventing DEC from overfitting to specific validation sets.
Second, excluding $\mathcal{L}_M$ also impairs generalization to the unseen TabPFN in ${\rm cf}_3$, evidenced by a substantial drop of 3.0\% on the Credit dataset. This confirms that $\mathcal{L}_M$ effectively aligns influence representations across diverse model structures.
Finally, the removal of both terms (w/o $\mathcal{L}_V, \mathcal{L}_M$) yields the lowest performance, demonstrating that our proposed invariant learning framework is critical for generalizable training data debugging.

\begin{table}[t]
  \centering
  \caption{Ablation Studies (debugging F1, \%) of  \methodname~under different configurations ($\rm cf_1$, $\rm cf_2$ and $\rm cf_3$) for influence vector computation. The DEC of \methodname~is trained on $\rm cf_1$, and evaluated on $\rm cf_2$ and $\rm cf_3$.}
  \label{tab:ablation_cf}
  \resizebox{\linewidth}{!}{ 
  \begin{tabular}{lcccccccccc}
    \toprule
    \multirow{2}{*}{\textbf{Method}} 
    & \multicolumn{3}{c}{\textbf{Adult}} 
    & \multicolumn{3}{c}{\textbf{Bank}} 
    & \multicolumn{3}{c}{\textbf{Credit}} \\
    \cmidrule(lr){2-4} \cmidrule(lr){5-7} \cmidrule(lr){8-10}
    & \textbf{${\rm cf}_1$} & \textbf{${\rm cf}_2$} & \textbf{${\rm cf}_3$} 
    & \textbf{${\rm cf}_1$} & \textbf{${\rm cf}_2$} & \textbf{${\rm cf}_3$} 
    & \textbf{${\rm cf}_1$} & \textbf{${\rm cf}_2$} & \textbf{${\rm cf}_3$} \\
    \midrule
    \methodname 
    & 87.23 & 86.92 & 86.07 & 85.65 & 85.17 & 85.22  & 83.03 & 82.54 & 83.21 \\
   \ \ w/o. $\mathcal{L}_{V}$
    & 86.53 & 83.49 & 82.02 & 85.03 & 82.15 & 82.27 & 82.28 & 79.36 & 80.03 \\
    \ \ w/o. $\mathcal{L}_{M}$
    & 87.02 & 84.74 & 82.46 & 84.83 & 84.47 & 83.13 & 81.76 & 80.43 & 78.26 \\
    \ \ w/o. $\mathcal{L}_{V}, \mathcal{L}_{M}$ 
    & 85.27 & 83.25 & 81.78 & 84.15 & 80.14 & 81.77 & 80.13 & 78.66 & 78.19 \\
    \bottomrule
  \end{tabular}
  }
\end{table}

\camera{
\subsection{Deployment Costs}
Table~\ref{tab:deployment} reports the deployment costs of \methodname~on the MovieLens dataset ($>$1.3M samples). \methodname~achieves comparable inference efficiency to DDA-repair with substantially higher debugging F1. Both methods require computing the influence values of training samples on validation samples; however, DDA-repair averages these values into a scalar, while DeMix retains them as a vector. DeMix's additional overhead relative to DDA-repair is the DEC inference pass ($\sim$3 min), which is negligible compared to the overall pipeline cost. For large-scale datasets, DeMix scales efficiently via validation sampling: using only 10\% of validation samples for influence computation reduces both runtime (from 35.4 to 6.6 min) and peak GPU memory (from 22.6 to 9.2 GB), while incurring a negligible drop in debugging F1 ($\sim$0.2\%).
}

\section{Conclusion}
In this paper, we introduce \methodname, a novel framework that leverages influence vectors for debugging training data with mixed error types. Our key insight is that different error types induce distinct patterns in model predictions across the validation set, which are effectively captured by influence vectors. Building on this insight, we formulate training data debugging as a multi-label classification problem that takes influence vectors as input features. We design a classifier to predict error types and introduce an intervention-based training strategy that ensures the classifier captures invariant and error-specific rationales, improving its generalization across different training setups.
Extensive experiments are conducted on 11 tasks spanning tabular data prediction, recommendation systems, and LLM alignment. The results demonstrate that \methodname~significantly outperforms state-of-the-art baselines, achieving substantial improvements in debugging F1 score and downstream task model performance after targeted repair. 

\begin{table}[t]
  \centering
  \caption{Deployment costs on MovieLens on a server with NVIDIA RTX 4090 GPU.}
  \label{tab:deployment}
  \resizebox{\linewidth}{!}{
  \begin{tabular}{lccc}
    \toprule
    \textbf{Method} & \textbf{Runtime (min)} & \textbf{Peak GPU Mem (GB)} & \textbf{Debugging F1 (\%)} \\
    \midrule
    DDA-repair & 32.1 & 14.4 & 60.3 \\
    \methodname & 35.4 & 22.6 & {82.1} \\
    \methodname~(10\% val.) & 6.6 & 9.2 & 81.9 \\
    \bottomrule
  \end{tabular}
  }
\end{table}

\begin{acks}
\camera{
This work is supported by the National Key Research and Development Program of China (No. 2024YFB4505203), National Natural Science Foundation of China (No. 62522211), and Key Research and Development Program of Xinjiang Uygur Autonomous Region (Grant No. 2023B01027, 2023B01027-1).
}
\end{acks}

\bibliographystyle{ACM-Reference-Format}
\balance
\bibliography{reference}

@inproceedings{influence_ICML17,
  title={Understanding black-box predictions via influence functions},
  author={Koh, Pang Wei and Liang, Percy},
  booktitle={International conference on machine learning},
  pages={1885--1894},
  year={2017},
  organization={PMLR}
}

@inproceedings{Dataset_regeneration_KDD24,
  title={Dataset regeneration for sequential recommendation},
  author={Yin, Mingjia and Wang, Hao and Guo, Wei and Liu, Yong and Zhang, Suojuan and Zhao, Sirui and Lian, Defu and Chen, Enhong},
  booktitle={Proceedings of the 30th ACM SIGKDD Conference on Knowledge Discovery and Data Mining},
  pages={3954--3965},
  year={2024}
}

@inproceedings{Shapley_data_pruning_KDD25,
  title={Shapley value-driven data pruning for recommender systems},
  author={Zhang, Yansen and Zhang, Xiaokun and Cui, Ziqiang and Ma, Chen},
  booktitle={Proceedings of the 31st ACM SIGKDD Conference on Knowledge Discovery and Data Mining V. 2},
  pages={3879--3888},
  year={2025}
}

@inproceedings{Shapley_Recsys25,
  title={Scalable Data Debugging for Neighborhood-based Recommendation with Data Shapley Values},
  author={Kersbergen, Barrie and Sprangers, Olivier and Karla{\v{s}}, Bojan and de Rijke, Maarten and Schelter, Sebastian},
  booktitle={Proceedings of the Nineteenth ACM Conference on Recommender Systems},
  pages={441--450},
  year={2025}
}

@inproceedings{LESS_ICML24,
  title={LESS: Selecting Influential Data for Targeted Instruction Tuning},
  author={Xia, Mengzhou and Malladi, Sadhika and Gururangan, Suchin and Arora, Sanjeev and Chen, Danqi},
  booktitle={International Conference on Machine Learning},
  pages={54104--54132},
  year={2024},
  organization={PMLR}
}

@inproceedings{DataMan_ICLR25,
  title={DataMan: Data Manager for Pre-training Large Language Models},
  author={Peng, Ru and Yang, Kexin and Zeng, Yawen and Lin, Junyang and Liu, Dayiheng and Zhao, Junbo},
  booktitle={The Thirteenth International Conference on Learning Representations}
}

@article{Data_Attribution_Survey25,
  title={A Survey of Data Attribution: Methods, Applications, and Evaluation in the Era of Generative AI},
  author={Deng, Junwei and Hu, Yuzheng and Hu, Pingbang and Li, Ting-Wei and Liu, Shixuan and Wang, Jiachen T and Ley, Dan and Dai, Qirun and Huang, Benhao and Huang, Jin and others},
  year={2025}
}

@inproceedings{Influence_Relabeling_ICLR21,
  title={Resolving training biases via influence-based data relabeling},
  author={Kong, Shuming and Shen, Yanyan and Huang, Linpeng},
  booktitle={International Conference on Learning Representations},
  year={2021}
}

@article{Disdetect_VLDB24,
  title={MisDetect: Iterative Mislabel Detection using Early Loss},
  author={Deng, Yuhao and Chai, Chengliang and Cao, Lei and Tang, Nan and Wang, Jiayi and Fan, Ju and Yuan, Ye and Wang, Guoren},
  journal={Proceedings of the VLDB Endowment},
  volume={17},
  number={6},
  pages={1159--1172},
  year={2024},
  publisher={VLDB Endowment}
}

@article{UniClean_VLDB25,
  title={UniClean: A Scalable Data Cleaning Solution for Mixed Errors based on Unified Cleaners and Optimized Cleaning Workflow},
  author={Ding, Xiaoou and Qian, Zekai and Wang, Hongzhi and Chen, Siying and Tang, Yafeng and Su, Hongbin and Hu, Huan and Wang, Chen},
  journal={Proceedings of the VLDB Endowment},
  volume={18},
  number={11},
  pages={4117--4130},
  year={2025},
  publisher={VLDB Endowment}
}

@inproceedings{Data_Glitches_KDD25,
  title={Data Glitches Discovery using Influence-based Model Explanations},
  author={Myrtakis, Nikolaos and Tsamardinos, Ioannis and Christophides, Vassilis},
  booktitle={Proceedings of the 31st ACM SIGKDD Conference on Knowledge Discovery and Data Mining V. 1},
  pages={1068--1079},
  year={2025}
}

@inproceedings{Spurious_Correlation_KDD25,
  title={Improving group robustness on spurious correlation via evidential alignment},
  author={Ye, Wenqian and Zheng, Guangtao and Zhang, Aidong},
  booktitle={Proceedings of the 31st ACM SIGKDD Conference on Knowledge Discovery and Data Mining V. 2},
  pages={3610--3621},
  year={2025}
}

@inproceedings{Discover_and_Cure_ICML23,
  title={Discover and cure: Concept-aware mitigation of spurious correlation},
  author={Wu, Shirley and Yuksekgonul, Mert and Zhang, Linjun and Zou, James},
  booktitle={International Conference on Machine Learning},
  pages={37765--37786},
  year={2023},
  organization={PMLR}
}

@inproceedings{ODIN_ICML24,
  title={ODIN: Disentangled Reward Mitigates Hacking in RLHF},
  author={Chen, Lichang and Zhu, Chen and Chen, Jiuhai and Soselia, Davit and Zhou, Tianyi and Goldstein, Tom and Huang, Heng and Shoeybi, Mohammad and Catanzaro, Bryan},
  booktitle={International Conference on Machine Learning},
  pages={7935--7952},
  year={2024},
  organization={PMLR}
}

@article{Trustworthy_AI_Data_NatMachIntell22,
  title={Advances, challenges and opportunities in creating data for trustworthy AI},
  author={Liang, Weixin and Tadesse, Girmaw Abebe and Ho, Daniel and Fei-Fei, Li and Zaharia, Matei and Zhang, Ce and Zou, James},
  journal={Nature Machine Intelligence},
  volume={4},
  number={8},
  pages={669--677},
  year={2022},
  publisher={Nature Publishing Group UK London}
}

@article{SAGA_SIGMOD23,
  title={SAGA: A scalable framework for optimizing data cleaning pipelines for machine learning applications},
  author={Siddiqi, Shafaq and Kern, Roman and Boehm, Matthias},
  journal={Proceedings of the ACM on Management of Data},
  volume={1},
  number={3},
  pages={1--26},
  year={2023},
  publisher={ACM New York, NY, USA}
}

@inproceedings{Data_Cleaning_Survey_SIGMOD16,
  title={Data cleaning: Overview and emerging challenges},
  author={Chu, Xu and Ilyas, Ihab F and Krishnan, Sanjay and Wang, Jiannan},
  booktitle={Proceedings of the 2016 international conference on management of data},
  pages={2201--2206},
  year={2016}
}

@inproceedings{Rock_Data_Cleaning_SIGMOD24,
  title={Rock: Cleaning Data by Embedding ML in Logic Rules},
  author={Bao, Xianchun and Bao, Zian and Binbin, Bie and Duan, QingSong and Fan, Wenfei and Lei, Hui and Li, Daji and Lin, Wei and Liu, Peng and Lv, Zhicong and others},
  booktitle={Companion of the 2024 International Conference on Management of Data},
  pages={106--119},
  year={2024}
}

@article{Opendataval_NIPS23,
  title={Opendataval: a unified benchmark for data valuation},
  author={Jiang, Kevin and Liang, Weixin and Zou, James Y and Kwon, Yongchan},
  journal={Advances in Neural Information Processing Systems},
  volume={36},
  pages={28624--28647},
  year={2023}
}

@article{Influence_embedding_NIPS23,
  title={Error discovery by clustering influence embeddings},
  author={Wang, Fulton and Adebayo, Julius and Tan, Sarah and Garcia-Olano, Diego and Kokhlikyan, Narine},
  journal={Advances in Neural Information Processing Systems},
  volume={36},
  pages={41765--41777},
  year={2023}
}

@article{Influence_Analysis_Survey_24,
  title={Training data influence analysis and estimation: A survey},
  author={Hammoudeh, Zayd and Lowd, Daniel},
  journal={Machine Learning},
  volume={113},
  number={5},
  pages={2351--2403},
  year={2024},
  publisher={Springer}
}

@inproceedings{Set_transformer,
  title={Set transformer: A framework for attention-based permutation-invariant neural networks},
  author={Lee, Juho and Lee, Yoonho and Kim, Jungtaek and Kosiorek, Adam and Choi, Seungjin and Teh, Yee Whye},
  booktitle={International conference on machine learning},
  pages={3744--3753},
  year={2019},
  organization={PMLR}
}

@inproceedings{DIEN,
  title={Deep interest evolution network for click-through rate prediction},
  author={Zhou, Guorui and Mou, Na and Fan, Ying and Pi, Qi and Bian, Weijie and Zhou, Chang and Zhu, Xiaoqiang and Gai, Kun},
  booktitle={Proceedings of the AAAI conference on artificial intelligence},
  volume={33},
  number={01},
  pages={5941--5948},
  year={2019}
}

@inproceedings{Influence_context_selection_NIPS25,
  title={Influence Guided Context Selection for Effective Retrieval-Augmented Generation},
  author={Deng, Jiale and Shen, Yanyan and Pei, Ziyuan and Chen, Youmin and Huang, Linpeng},
  booktitle={The Thirty-ninth Annual Conference on Neural Information Processing Systems}
}

@inproceedings{JTT_ICML21,
  title={Just train twice: Improving group robustness without training group information},
  author={Liu, Evan Z and Haghgoo, Behzad and Chen, Annie S and Raghunathan, Aditi and Koh, Pang Wei and Sagawa, Shiori and Liang, Percy and Finn, Chelsea},
  booktitle={International Conference on Machine Learning},
  pages={6781--6792},
  year={2021},
  organization={PMLR}
}

@inproceedings{Spurious_tabular_ICML25,
  title={Latent Score-Based Reweighting for Robust Classification on Imbalanced Tabular Data},
  author={Tong, Yunze and Zhang, Fengda and Tang, Zihao and Gao, Kaifeng and Huang, Kai and Lyu, Pengfei and Xiao, Jun and Kuang, Kun},
  booktitle={Forty-second International Conference on Machine Learning}
}

@article{Grass_influence_NIPS25,
  title={GraSS: Scalable Influence Function with Sparse Gradient Compression},
  author={Hu, Pingbang and Melkonian, Joseph and Tang, Weijing and Zhao, Han and Ma, Jiaqi W},
  journal={arXiv preprint arXiv:2505.18976},
  year={2025}
}

@article{Influence_GPT_NIPS25,
  title={What is your data worth to gpt? llm-scale data valuation with influence functions},
  author={Choe, Sang Keun and Ahn, Hwijeen and Bae, Juhan and Zhao, Kewen and Kang, Minsoo and Chung, Youngseog and Pratapa, Adithya and Neiswanger, Willie and Strubell, Emma and Mitamura, Teruko and others},
  journal={arXiv preprint arXiv:2405.13954},
  year={2024}
}

@inproceedings{Shapley_negative_clustering_ICML24,
  title={Exploiting negative samples: a catalyst for cohort discovery in healthcare analytics},
  author={Zheng, Kaiping and Chua, Horng-Ruey and Herschel, Melanie and Jagadish, HV and Ooi, Beng Chin and Yip, James Wei Luen},
  booktitle={Forty-first International Conference on Machine Learning},
  year={2024}
}

@inproceedings{GSAT_ICML22,
  title={Interpretable and generalizable graph learning via stochastic attention mechanism},
  author={Miao, Siqi and Liu, Mia and Li, Pan},
  booktitle={International conference on machine learning},
  pages={15524--15543},
  year={2022},
  organization={PMLR}
}

@inproceedings{DeepVIB_ICLR17,
  title={Deep Variational Information Bottleneck},
  author={Alemi, Alexander A and Fischer, Ian and Dillon, Joshua V and Murphy, Kevin},
  booktitle={International Conference on Learning Representations},
  year={2017}
}

@article{InfoNCE,
  title={Representation learning with contrastive predictive coding},
  author={Oord, Aaron van den and Li, Yazhe and Vinyals, Oriol},
  journal={arXiv preprint arXiv:1807.03748},
  year={2018}
}

@inproceedings{TabPFN,
  title={TabPFN: A Transformer That Solves Small Tabular Classification Problems in a Second},
  author={Hollmann, Noah and M{\"u}ller, Samuel and Eggensperger, Katharina and Hutter, Frank},
  booktitle={The Eleventh International Conference on Learning Representations}
}

@inproceedings{DIN,
  title={Deep interest network for click-through rate prediction},
  author={Zhou, Guorui and Zhu, Xiaoqiang and Song, Chenru and Fan, Ying and Zhu, Han and Ma, Xiao and Yan, Yanghui and Jin, Junqi and Li, Han and Gai, Kun},
  booktitle={Proceedings of the 24th ACM SIGKDD international conference on knowledge discovery \& data mining},
  pages={1059--1068},
  year={2018}
}

@article{DHEN,
  title={DHEN: A deep and hierarchical ensemble network for large-scale click-through rate prediction},
  author={Zhang, Buyun and Luo, Liang and Liu, Xi and Li, Jay and Chen, Zeliang and Zhang, Weilin and Wei, Xiaohan and Hao, Yuchen and Tsang, Michael and Wang, Wenjun and others},
  journal={arXiv preprint arXiv:2203.11014},
  year={2022}
}

@misc{qwen2.5,
    title = {Qwen2.5: A Party of Foundation Models},
    url = {https://qwenlm.github.io/blog/qwen2.5/},
    author = {Qwen Team},
    month = {September},
    year = {2024}
}

@article{DPO,
  title={Direct preference optimization: Your language model is secretly a reward model},
  author={Rafailov, Rafael and Sharma, Archit and Mitchell, Eric and Manning, Christopher D and Ermon, Stefano and Finn, Chelsea},
  journal={Advances in neural information processing systems},
  volume={36},
  pages={53728--53741},
  year={2023}
}

@article{Information_bottleneck,
  title={The information bottleneck method},
  author={Tishby, Naftali and Pereira, Fernando C and Bialek, William},
  journal={arXiv preprint physics/0004057},
  year={2000}
}

@inproceedings{chen2024odin,
  title={ODIN: Disentangled Reward Mitigates Hacking in RLHF},
  author={Chen, Lichang and Zhu, Chen and Chen, Jiuhai and Soselia, Davit and Zhou, Tianyi and Goldstein, Tom and Huang, Heng and Shoeybi, Mohammad and Catanzaro, Bryan},
  booktitle={International Conference on Machine Learning},
  pages={7935--7952},
  year={2024},
  organization={PMLR}
}

@inproceedings{From_list_to_emoji_ACL25,
  title={From lists to emojis: How format bias affects model alignment},
  author={Zhang, Xuanchang and Xiong, Wei and Chen, Lichang and Zhou, Tianyi and Huang, Heng and Zhang, Tong},
  booktitle={Proceedings of the 63rd Annual Meeting of the Association for Computational Linguistics (Volume 1: Long Papers)},
  pages={26940--26961},
  year={2025}
}

@inproceedings{Cleanlab_label,
    title={Model-agnostic label quality scoring to detect real-world label errors},
    author={Kuan, Johnson and Mueller, Jonas},
    booktitle={ICML DataPerf Workshop},
    year={2022}
}

@inproceedings{Cleanlab_ood,
    title={Back to the Basics: Revisiting Out-of-Distribution Detection Baselines},
    author={Kuan, Johnson and Mueller, Jonas},
    booktitle={ICML Workshop on Principles of Distribution Shift},
    year={2022}
}

@inproceedings{
    DIR,
    title={Discovering Invariant Rationales for Graph Neural Networks},
    author={Ying-Xin Wu and Xiang Wang and An Zhang and Xiangnan He and Tat-seng Chua},
    booktitle={ICLR},
    year={2022},
}

@article{Imbalance_FCS,
  title={A comprehensive survey on imbalanced data learning},
  author={Gao, Xinyi and Xie, Dongting and Zhang, Yihang and Wang, Zhengren and Chen, Chong and He, Conghui and Yin, Hongzhi and Zhang, Wentao},
  journal={Frontiers of Computer Science},
  volume={20},
  number={11},
  pages={2011622},
  year={2026},
  publisher={Springer}
}

@article{Alignment_FCS,
  title={The gains do not make up for the losses: a comprehensive evaluation for safety alignment of large language models via machine unlearning},
  author={Zhao, Weixiang and Hu, Yulin and Sui, Xingyu and Li, Zhuojun and Deng, Yang and Zhao, Yanyan and Qin, Bing and Che, Wanxiang},
  journal={Frontiers of Computer Science},
  volume={20},
  number={2},
  pages={2002319},
  year={2026},
  publisher={Springer}
}

@article{Data_quality_FCS,
  title={Data preparation and quality for code-centric generative software engineering tasks: a systematic literature review},
  author={Weng, Shihao and Feng, Yang and Yin, Yining and Zhang, Zhenlun and Xu, Baowen},
  journal={Frontiers of Computer Science},
  volume={20},
  number={9},
  pages={2009203},
  year={2026},
  publisher={Springer}
}

\appendix

\section*{\LARGE Appendix}

\begin{figure}[t]
    \centering
    \includegraphics[width=1\linewidth]{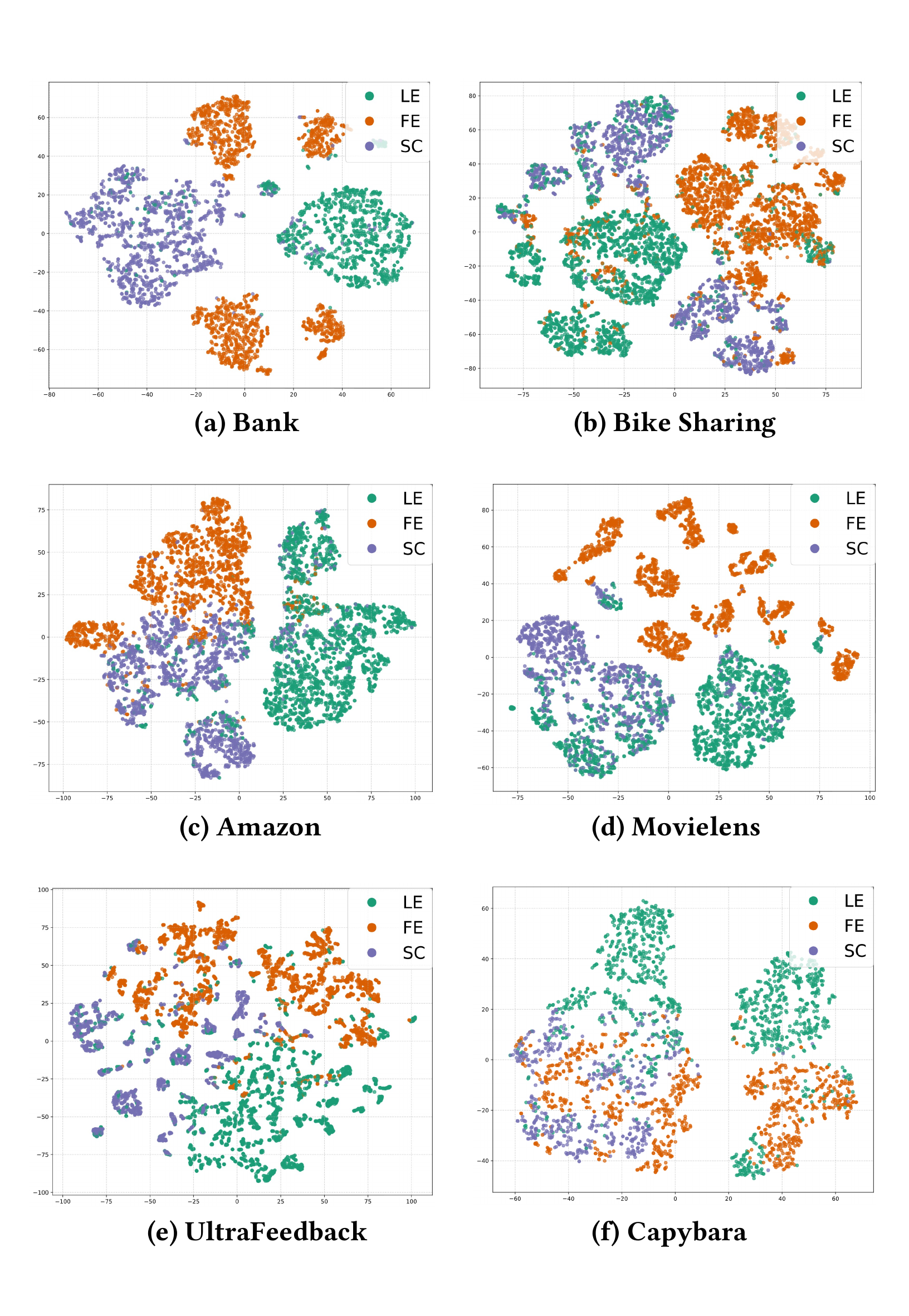}
    \caption{t-SNE visualization of influence vectors of erroneous samples, colored by error types.}
    \label{fig:visual_app}
\end{figure}

\section{Visualizations of Influence Vectors} 
\label{app:visual}
Figure~\ref{fig:visual_app} shows t-SNE visualizations of influence vectors across diverse tasks. Samples with the same error type consistently form cohesive clusters, while different error types are well separated, demonstrating the effectiveness of influence vectors in characterizing error patterns across tabular, recommendation, and LLM alignment tasks. 
For LLM alignment, the separation is less distinct, with FEs often overlapping with other error types. This is because LLMs are relatively insensitive to token-level perturbations; in their massive parameter space, such subtle text-noise signals are easily diluted, causing FEs to entangle with other error categories in the projected space.

\section{Case Study} 
\label{app: case}
\begin{figure}[t]
    \centering
    \includegraphics[width=1\linewidth]{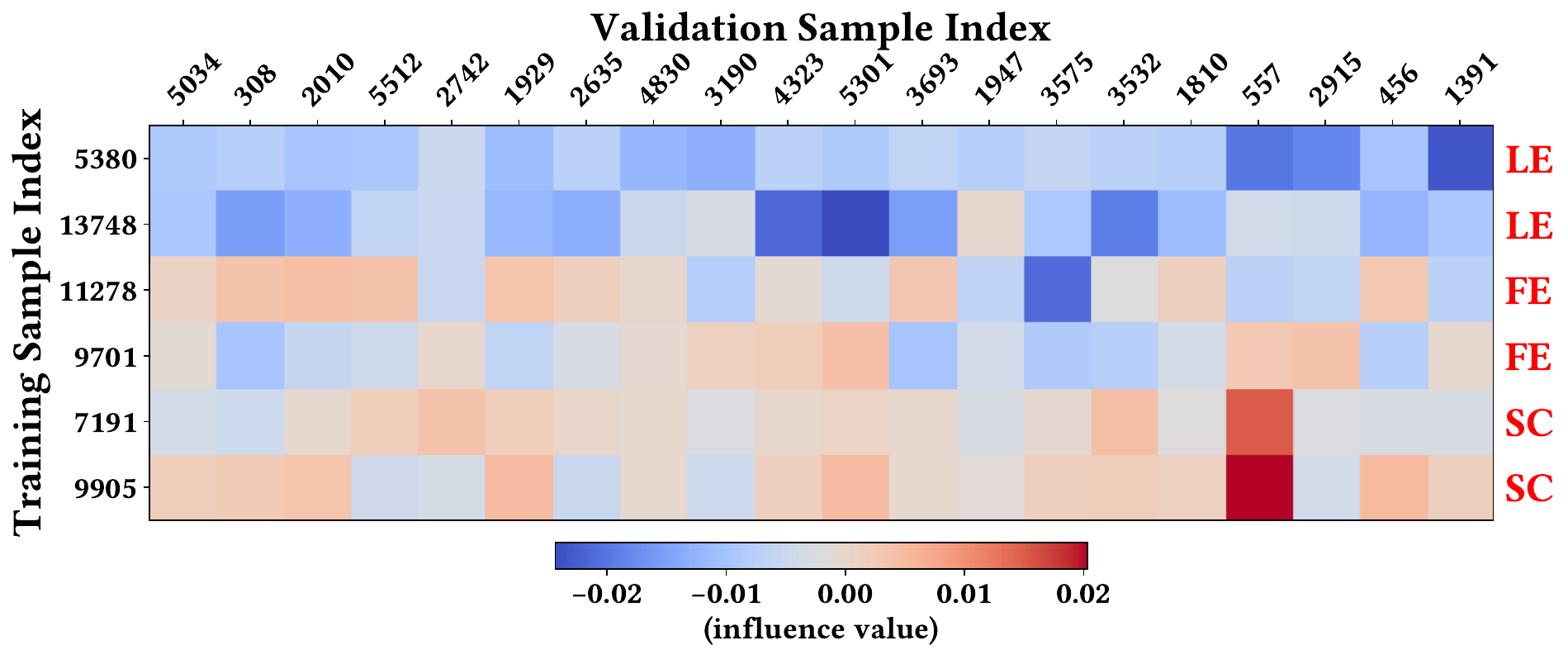}
    \caption{A case study on the Adult dataset, where blue and red regions indicate negative and positive influence values, respectively.}
    \label{fig:case}
\end{figure}

Figure~\ref{fig:case} visualizes the influence vectors of six training samples containing different error types across 20 validation instances from the Adult dataset. The heatmap reveals that distinct error types manifest as unique patterns in the influence vectors. Specifically, LEs exhibit consistently negative influence (indicated by blue regions in rows 1 and 2), reflecting a more detrimental effect on the prediction of similar validation samples (e.g., sample 5380 and sample 1391 are similar). This suggests that LEs are particularly harmful, likely inducing significant shifts in the decision boundary. Conversely, influence values of FEs remain close to zero (rows 3 and 4), implying a negligible impact due to the stochastic nature of random feature corruption. Notably, SCs display sharp positive spikes on specific validation samples sharing the same minority group. For instance, training samples 7191 and 9905 strongly support validation sample 557 (dark red), as they share the minority attribute combination (\textit{Female}, \textit{$>$50K}). These distinguishable patterns empirically validate that influence vectors effectively encode error-specific semantics, a property our framework leverages for accurate data debugging.

\begin{figure*}[h]
    \centering
    \includegraphics[width=0.88\linewidth]{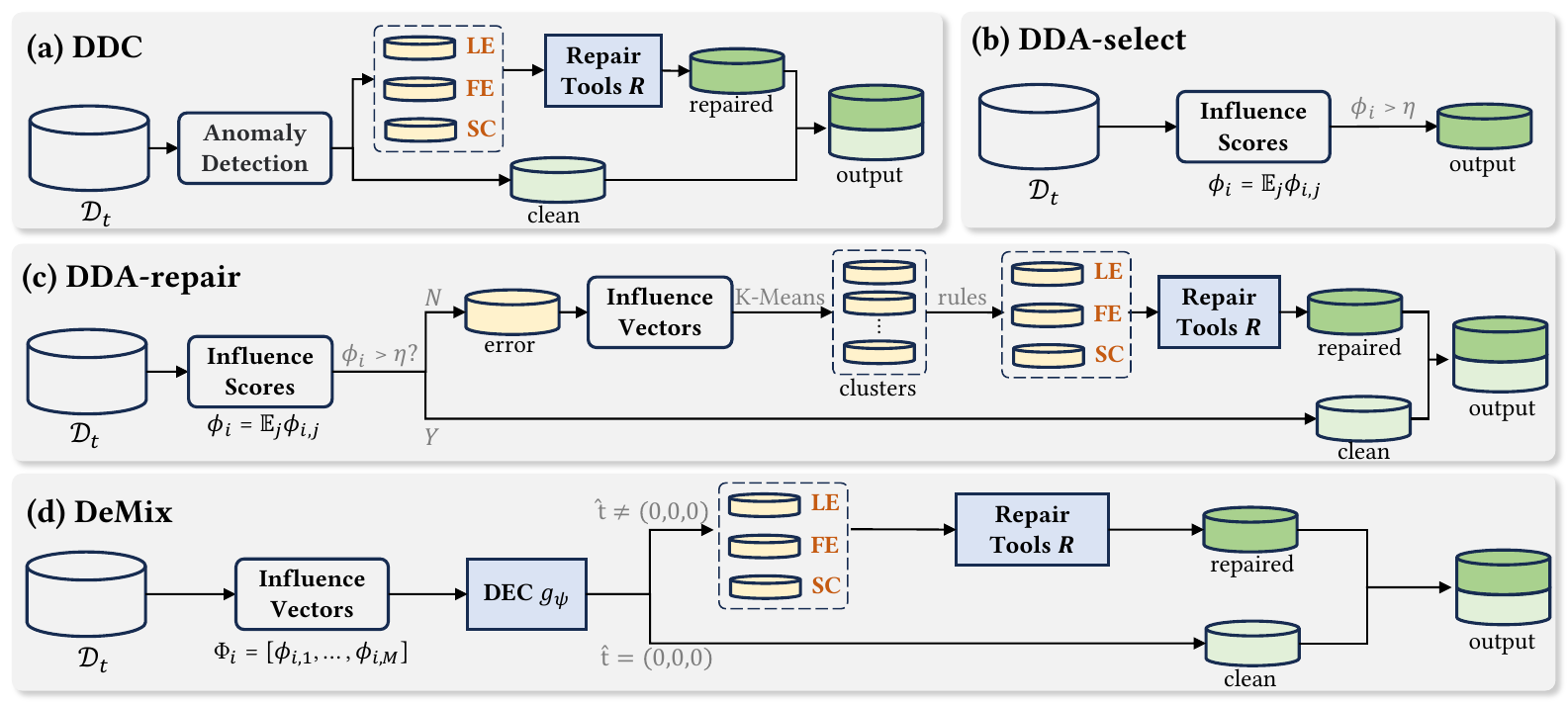}
    \caption{Overall workflow of baselines.}
    \label{fig:workflow}
\end{figure*}

\section{\camera{Theoretical Analysis}}
\label{app:theory}
\camera{The training objective of DEC is grounded in the Information Bottleneck (IB) principle~\cite{Information_bottleneck, DeepVIB_ICLR17, GSAT_ICML22}. Let $\Phi$ denote influence vectors, $E$ denote the error-type variable, and $H=q_{\psi}(\Phi)$ denote the representation produced by the Set Transformer encoder. Since $\Phi$ inevitably contains configuration-specific noise introduced by the validation set and task model used for influence computation, DEC aims to learn a minimal sufficient representation:
\begin{equation}
    H^*=\arg\max_H \left[I(H;E)-\beta\, I(H;\Phi)\right],
    \label{eq:ib_objective}
\end{equation}
where $I(\cdot;\cdot)$ denotes mutual information. The first term ensures that $H$ preserves error-specific semantics from $\Phi$, while the second term compresses away irrelevant configuration noise. Below we derive tractable surrogates for each term.
\paragraph{Maximizing $I(H;E)$.}
The true posterior $p(E|H)$ is intractable, so we introduce the DEC prediction head $q_{\omega}(E|H)$ as a variational approximation~\cite{DeepVIB_ICLR17, GSAT_ICML22}:
\begin{equation}
    \begin{aligned}
    I(H;E)
    &= \mathbb{E}_{H,E}\left[\log \frac{p(E|H)}{p(E)}\right] \\
    &= \mathbb{E}_{H,E}\left[\log q_{\omega}(E|H)\right]-\mathbb{E}_{E}\left[\log p(E)\right] \\
    &\quad +\mathbb{E}_{H}\left[{\rm KL}\left(p(E|H)\|q_{\omega}(E|H)\right)\right] \\
    &\ge \mathbb{E}_{H,E}\left[\log q_{\omega}(E|H)\right]-\mathbb{E}_{E}\left[\log p(E)\right].
    \end{aligned}
    \label{eq:app_pred_bound}
\end{equation}
Since $-\mathbb{E}_{E}[\log p(E)]$ is constant during training, maximizing this lower bound reduces to maximizing $\mathbb{E}[\log q_{\omega}(E|H)]$. Because each error type is modeled as an independent Bernoulli variable, this is equivalent to minimizing the element-wise BCE loss $\mathcal{L}_{\rm pred}$.
\paragraph{Minimizing $I(H;\Phi)$.}
The compression term $I(H;\Phi)$ aims to discard the information in $\Phi$ that is irrelevant to the error semantics $E$. In our setting, the dominant source of such nuisance information is the configuration $C=(V,M)$ used to compute influence vectors, where $V$ denotes the validation subset and $M$ denotes the task model. Changing either $V$ or $M$ alters the numerical scale, sparsity, and local patterns of $\Phi$, even when the underlying training sample and its error type remain unchanged. Therefore, after $\mathcal{L}_{\rm pred}$ preserves the error-specific semantics in $H$, we focus on suppressing the conditional configuration information $I(H;C|E)$ as the configuration-related component of the compression objective. By the chain rule of mutual information,
\begin{equation}
    I(H;E,C)=I(H;E)+I(H;C|E),
    \label{eq:app_chain_rule}
\end{equation}
so minimizing the configuration-specific part amounts to minimizing $I(H;C|E)$. This conditional mutual information admits the exact KL form:
\begin{equation}
    I(H;C|E)
    =
    \mathbb{E}_{e,c}\left[
    {\rm KL}\left(p(H|E=e,C=c)\,\|\,p(H|E=e)\right)
    \right].
    \label{eq:app_cond_mi}
\end{equation}
$I(H;C|E)=0$ if and only if $p(H|E=e,C=c)=p(H|E=e)$ for all configurations $c$, i.e., the representation is invariant to configurations within each error type. Since $C=(V,M)$, we decompose:
\begin{equation}
    I(H;C|E)=I(H;V|E)+I(H;M|E,V).
    \label{eq:app_config_decomp}
\end{equation}
\paragraph{Connection to $\mathcal{L}_V$ and $\mathcal{L}_M$.}
The validation-invariant loss $\mathcal{L}_V$ minimizes an empirical surrogate of $I(H;V|E)$ by pulling together representations of the same sample computed from different validation subsets, while its contrastive negatives avoid representation collapse across different error types. The model-invariant loss $\mathcal{L}_M$ minimizes an empirical surrogate of $I(H;M|E,V)$ by directly aligning representations of the same sample computed from different task models. Under the common local Gaussian approximation $p(H|E=e,C=c)=\mathcal{N}(\mu_{e,c},\sigma^2 I)$, these KL terms are proportional to squared distances between configuration-specific representation means, which justifies the pairwise alignment forms used in $\mathcal{L}_V$ and $\mathcal{L}_M$. Therefore, DEC optimizes the IB objective by using $\mathcal{L}_{\rm pred}$ to preserve $I(H;E)$ and using $\mathcal{L}_V,\mathcal{L}_M$ to suppress the configuration component of $I(H;\Phi)$.}

\section{Baseline Workflows}
\label{app:workflow}

As illustrated in Figure~\ref{fig:workflow}, we compare \methodname~against three representative baselines.
(a) DDC treats erroneous data as anomalies and applies type-specific repair based on detection results.
(b) DDA-select aggregates influence values over validation data into scalar scores and discards samples below a threshold $\eta$.
(c) DDA-repair extends DDA-select with a heuristic repair pipeline: it selects detrimental samples, clusters their influence vectors with K-Means, and assigns error types by comparing cluster centers with reference vectors generated from a clean calibration dataset with injected errors.
(d) \methodname~simplifies this pipeline by using a parameterized DEC $g_\psi$ to directly map each influence vector $\Phi_i$ to its error configuration $\hat{t}$, enabling simultaneous diagnosis and type-specific repair.

\section{Hyperparameters}
\label{app:hyperparams}
\camera{
We conduct a systematic grid search to optimize the two invariance regularization weights $\lambda_1$ and $\lambda_2$ in Eq.~\eqref{eq:overall loss}. Table~\ref{tab:hyperparam} reports the debugging F1 (\%) on the Adult dataset with $\alpha=0.5$. Setting $\lambda_1=\lambda_2=0.1$ yields the best performance. Higher regularization (e.g., 1) over-prioritizes invariance at the expense of prediction accuracy, while lower values (e.g., 0.01) diminish the information bottleneck benefits by retaining more configuration-specific noise.
}

\begin{table}[t]
  \small
  \centering
  \caption{Grid search over $\lambda_1$ and $\lambda_2$ on Adult ($\alpha=0.5$). Reported metric: debugging F1-score (\%).}
  \label{tab:hyperparam}
  \begin{tabular}{ccccc}
    \toprule
     & $\lambda_1=0.01$ & $\lambda_1=0.1$ & $\lambda_1=0.5$ & $\lambda_1=1$ \\
    \midrule
    $\lambda_2=0.01$ & 85.07 & 84.98 & 85.19 & 84.93 \\
    $\lambda_2=0.1$  & 85.26 & \textbf{85.32} & 85.22 & 84.91 \\
    $\lambda_2=0.5$  & 85.17 & 85.29 & 85.15 & 85.03 \\
    $\lambda_2=1$    & 85.08 & 84.95 & 85.01 & 84.82 \\
    \bottomrule
  \end{tabular}
\end{table}

\end{document}